\documentclass[10pt,twocolumn,letterpaper]{article}

\usepackage[pagenumbers]{wacv}
\usepackage{graphicx}
\usepackage{amsmath,amssymb}
\usepackage{booktabs}
\usepackage{subcaption}
\usepackage{multirow}
\usepackage{float}
\usepackage{algorithm}
\usepackage{algpseudocode}
\usepackage{pifont}
\usepackage{colortbl}
\newcommand{\cmark}{\textcolor{green!60!black}{\ding{51}}}
\newcommand{\xmark}{\textcolor{red!70!black}{\ding{55}}}
\usepackage{xspace}

\usepackage{placeins}

\definecolor{wacvblue}{rgb}{0.21,0.49,0.74}
\usepackage[breaklinks,colorlinks,allcolors=wacvblue]{hyperref}
\usepackage[capitalize]{cleveref}

\newcommand{\mavupapertitle}{EARS: Evidence-Aware Retrieval via Sound and Motion%
\texorpdfstring{\\}{ }for Budgeted Video Reasoning}

\title{\mavupapertitle}
\author{Shih-Yao Lin\\
Independent Researcher\\
{\tt\small mikeslin.research@gmail.com}}

\hypersetup{
  pdftitle={EARS: Evidence-Aware Retrieval via Sound and Motion for Budgeted Video Reasoning},
  pdfauthor={Shih-Yao Lin}
}

\begin{document}
\maketitle
\begin{abstract}
Uniform frame sampling can miss brief evidence in sparse-event videos when
the VLM sees only $K$ frames. We propose \textbf{EARS} (Evidence-Aware
Retrieval via Sound and motion), a training-free, query-free front-end that
ranks frames using optical-flow motion and audio energy, then applies temporal
non-maximum suppression for diversity. The evidence score alone has frame-level
AP $0.39$; used only to select frames for the VLM, EARS with temporal diversity
reaches localization AP $0.94$, compared with $0.88$ under Uniform sampling.
In the updated matched study, at $K\!=\!1$ EARS reaches XD-subset F1
$0.660$ versus $0.614$ for centered Uniform (the paired interval includes
zero); earlier backbone studies use separate protocols. On
mostly-silent UCF-Crime, however, EARS slightly trails Uniform. A single
configuration is used across both benchmarks; the new controls do not
repeat the original held-out search. EARS therefore
targets low-budget videos whose salient evidence aligns with motion or audio;
on dense narratives or stealthy events, uniform or semantic selection remains
preferable, and cheap modality statistics cannot reliably route between them.
\textbf{Code:} \url{https://github.com/mikeslin-ir/EARS}
\end{abstract}

\begin{figure*}[t]
\centering
\includegraphics[width=0.72\textwidth]{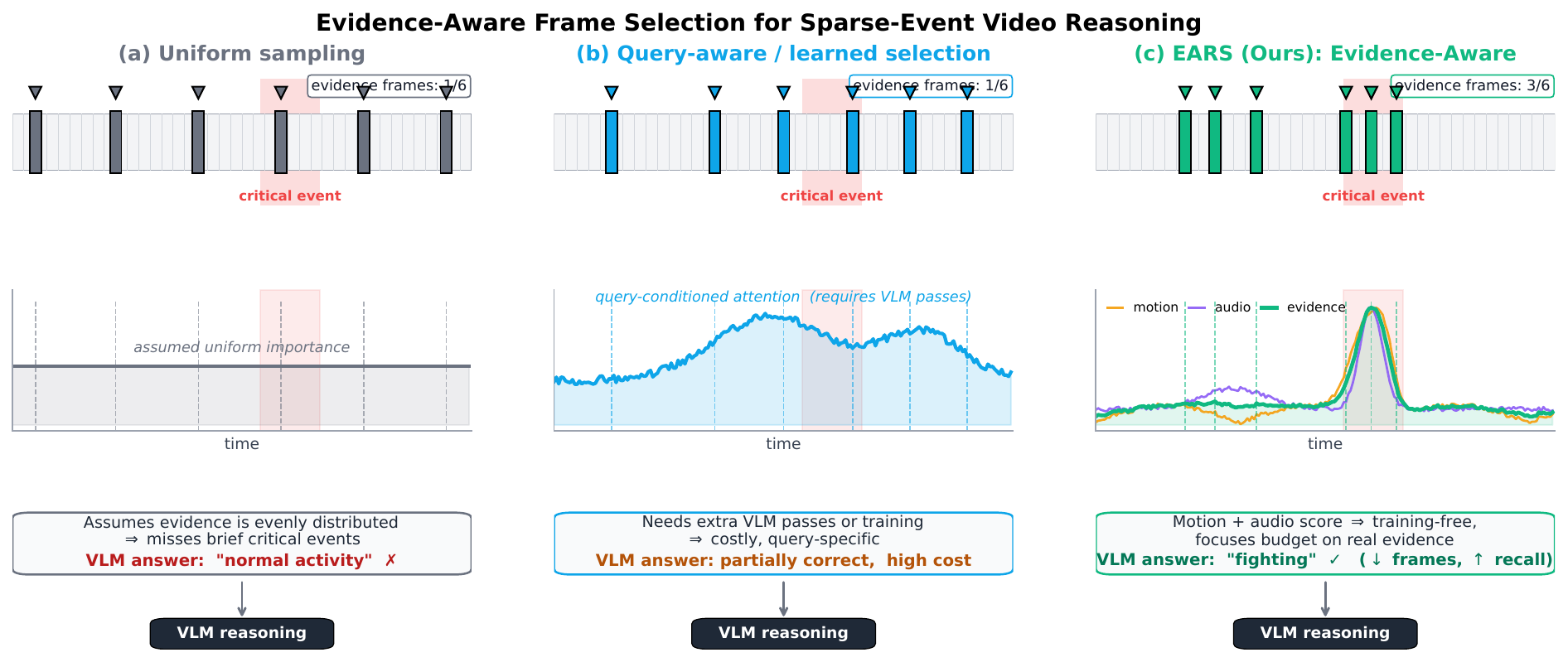}
\vspace{-5pt}
\caption{\textbf{Evidence-aware frame selection for sparse-event video reasoning.}
Under $K{=}6$, \textbf{(a)} Uniform largely misses a brief critical event;
\textbf{(b)} query-aware/learned selectors~\cite{sevila,videoagent,aks}
recover it but require a query, training, or extra model passes; \textbf{(c)}
EARS instead routes frames using training-free motion--audio evidence.}
\label{fig:teaser}
\end{figure*}

\section{Introduction}
\label{sec:intro}

Video understanding requires reasoning over long temporal sequences, where informative content is often sparse and unevenly distributed. Recent vision-language models (VLMs)~\cite{videochat,videollama,videollava,qwen2vl} have achieved strong performance on multimodal reasoning tasks, but their application to long videos remains computationally expensive, since VLM context capacity grows much slower than video length~\cite{longva,malmm}.

A common approach is to sample a fixed number of frames uniformly across time. This strategy implicitly assumes that useful information is evenly distributed. However, real-world videos violate this assumption. In many scenarios, such as surveillance and safety monitoring~\cite{ucfcrime,xdviolence}, important events occur only briefly, while the majority of frames contain little relevant information.
This mismatch leads to a fundamental inefficiency: processing more frames does not necessarily improve reasoning if those frames lack meaningful evidence. Under limited frame budgets, uniform sampling may miss critical evidence entirely~\cite{sevila,aks}.

In this work, we focus on \emph{sparse-event video reasoning} under \emph{tight frame budgets}, and argue that in this regime video reasoning should operate on \textbf{evidence rather than frames}. Instead of treating all frames equally, models should identify and prioritize temporally localized, information-rich regions.
To this end, we propose a training-free multimodal evidence selection front-end that leverages cheap, modality-intrinsic motion and audio signals to estimate a per-frame evidence score. These complementary modalities capture visual dynamics and acoustic activity, enabling detection of salient evidence without any learning or task-specific supervision (Sec.~\ref{sec:method}). In contrast to query-aware localizers~\cite{sevila,videoagent} or learned keyframe samplers~\cite{aks}, our method requires no additional VLM forward passes and no training, and is \emph{complementary} to such methods: it can be combined with semantic/query-aware signals or used as a cheap fallback when those signals are unavailable.

We instantiate this idea as \textbf{EARS} (Evidence-Aware Retrieval via Sound and motion), a lightweight, training-free front-end that selects frames from high-evidence regions and feeds them to a downstream VLM (Fig.~\ref{fig:ic}). Unlike query-aware or learned selectors~\cite{sevila,aks,videoagent}, EARS runs \emph{zero-shot per video}: no query, no auxiliary VLM passes, no learned scorer.
Experiments on XD-Violence~\cite{xdviolence} and UCF-Crime~\cite{ucfcrime} (Sec.~\ref{sec:exp}) show that EARS's gain is sharply conditional on the regime: it strongly improves \textbf{XD-Violence} when audio-visual evidence aligns with the anomaly, but slightly trails Uniform on mostly-silent, visually subtle \textbf{UCF-Crime}. On the dense-narrative \textbf{EgoSchema} benchmark uniform sampling is also stronger, confirming that EARS is most appropriate when evidence is sparse, temporally localized, and correlated with motion or audio --- not a universal frame selector.

\textbf{Contributions.}
\begin{itemize}
\setlength{\itemsep}{1pt}
\item \textbf{Evidence-centric framing.} We recast sparse-event video
reasoning as \emph{evidence selection} rather than uniform frame
sampling, isolating the regime where the mismatch matters: tight
budgets ($K\!\le\!4$) over temporally localized events.

\item \textbf{EARS, a training-free, query-free front-end.} A fused
motion+audio evidence score with an NMS-style temporal-diversity
operator (IC), controlled by a single coverage knob. EARS adds
\emph{no} VLM forward passes, no supervision, and no learned encoder, and
reduces to uniform sampling in a known limit. It runs one configuration
everywhere: the original held-out study is retained separately, while the
updated controls fix the same weights without a new search.

\item \textbf{Regime-conditional evaluation and scope.}
A $+0.047$ XD-subset F1 point difference at $K\!=\!1$ against centered
Uniform in the updated study; separate original studies of four VLM backbones spanning two families
and three generations (Qwen2/2.5/3-VL and cross-family LLaVA-NeXT-Video)
and query-aware selectors; lower UCF point estimates on the
mostly-silent UCF-Crime; and a
\emph{negative} result on dense-narrative EgoSchema. To our knowledge
this is the first work to systematically map \emph{when} training-free
frame selection helps a VLM and when it does not --- a missing
ingredient for practical deployment of frame-budgeted video reasoning.

\item \textbf{A boundary result on per-video routing.} If selection helps
on some videos and hurts on others, the obvious next step is to route
between them per video using the signals we have already computed. We show
this does not work: a per-video oracle would gain $+0.104$ over always
using EARS, yet a router built from eleven modality-intrinsic statistics
scores below never routing at all, and no statistic predicts the per-video
outcome once multiple comparisons are accounted for ($0$ of $92$ tests
survive correction). The headroom is real and the cheap route to it is
closed --- a constraint on a design space that recent adaptive-selection
work leaves unexamined.
\end{itemize}
\section{Related Work}
\label{sec:related}

\subsection{Video-Language Models}

Recent video-language models (VLMs) extend image-language pretraining to the
temporal domain by sampling a fixed number of frames from an input video and
feeding their visual tokens to a large language model (LLM).
Representative systems include \mbox{VideoChat}~\cite{videochat},
\mbox{Video-LLaMA}~\cite{videollama}, \mbox{Video-LLaVA}~\cite{videollava} and
\mbox{Qwen2-VL}~\cite{qwen2vl}, which now achieve strong performance on
short-form video QA and captioning.
For long-form videos, recent work either extrapolates the LLM context
length~\cite{longva} or designs memory mechanisms to summarize the full
sequence~\cite{malmm}, both of which still rely on uniformly sampled
frames as the underlying visual stream.
Standardized evaluations such as MVBench~\cite{videochat2} report that
performance degrades sharply when the frame budget is small, indicating that
\emph{which} frames are shown matters as much as how many.
We share this motivation but, in contrast to methods that scale up the
context window, focus on selecting a small, information-dense subset of
frames before any VLM computation is performed.

\subsection{Frame Sampling and Keyframe Selection}

Uniform sampling is the de-facto default in VLM pipelines because of its
simplicity, but it implicitly assumes information is evenly distributed in
time.
Earlier work on adaptive video recognition such as
ClipBERT~\cite{clipbert} and AdaFrame~\cite{adaframe} already showed that
sparse, content-aware sampling can match or beat dense sampling under tight
budgets.
For modern multimodal LLMs, SeViLA~\cite{sevila} chains a BLIP-2-based
\emph{localizer} and an \emph{answerer} to find query-relevant
keyframes; AKS~\cite{aks} formulates keyframe selection as an
optimization that jointly maximizes prompt relevance and temporal
coverage; M-LLM-FS~\cite{mllmfs} scores frames with a multimodal
LLM; and VideoTree~\cite{videotree} builds a hierarchical CLIP-based
frontier over the video.
Agent-style approaches such as VideoAgent~\cite{videoagent} iteratively
query a vision tool to gather evidence.
Recent training-free selectors reduce this overhead while retaining
semantic guidance. WFS-SB~\cite{wfssb2026} detects wavelet-smoothed
semantic boundaries before relevance-aware allocation;
EFS~\cite{efs2026} partitions videos with DINO features and selects a
query-relevant anchor per event; and MEC~\cite{mec2026} constructs one
query-conditioned evidence-to-context ranking reusable across budgets.
KeyScore~\cite{lin2025captions} combines caption similarity, temporal
representativeness, and contextual drop impact to score frames, together
with spatio-temporal adaptive clustering for compact, diverse proposals.
Complementing selection methods, STEC~\cite{lin2026stec} is a
reference-free diagnostic metric that evaluates a sampled frame set through
spatial entropy, temporal coverage, and non-redundancy without claiming to
predict downstream task accuracy.
These methods deliver clear gains, but all semantic selectors above
consume a task query and a learned image/video representation; several
also call an LLM/VLM during selection.
Our work is \emph{complementary} rather than competing: we use cheap,
modality-intrinsic signals (motion and audio energy) that require no
learning and no extra VLM forward passes, providing a training-free
front-end for the sparse-event, low-budget regime. The same selected
frames can be fed to any off-the-shelf VLM, and the evidence score is
additive with semantic/query-aware signals (e.g.\ CLIP/SigLIP
similarity) when those are available --- see Sec.~\ref{sec:exp_sota}.

\subsection{Efficient Video Understanding}

A parallel line of work reduces computation \emph{inside} the model rather
than at the input.
Token-level methods such as TokenLearner~\cite{tokenlearner} and Token
Merging~\cite{tome} dynamically prune or merge spatio-temporal tokens
during the forward pass.
These techniques are largely orthogonal to frame selection and can be
combined with our method: we reduce the number of frames \emph{before}
tokenization, while token pruning further compresses the remaining
representation.
We view our IC mechanism (Sec.~\ref{sec:method}) as an explicit, temporal
counterpart of token merging operating on the evidence axis.

A recent wave of audio-visual token-compression methods---%
OmniZip~\cite{omnizip2026}, EchoingPixels~\cite{echoingpixels2026}, and
OmniSIFT~\cite{omnisift2026}---prunes or merges multimodal tokens
\emph{after} the encoder to fit long videos into a fixed context.
These are \emph{post-encoder} operations and are orthogonal to, and
combinable with, EARS, which acts \emph{pre-encoder}: it decides which
frames enter the VLM at all, so no encoder pass is spent on frames that
are later discarded. On the selection side, BOLT~\cite{bolt2025}
similarly finds uniform sampling suboptimal for long video, but scores
frames by query--frame similarity, whereas our signal is query-free and
modality-intrinsic. Finally, our ``evidence'' framing connects to
temporally-grounded VLMs~\cite{factorized2025} and grounded multi-hop
VideoQA~\cite{mhvidqa2025}, which localize supporting moments for a
question; EARS instead surfaces salient moments \emph{without} a
question, as a cheap default when no query is available.

\subsection{Multimodal Event and Audio-Visual Analysis}

Audio and visual streams have long been recognized as complementary for
event and anomaly detection.
On the surveillance side, UCF-Crime~\cite{ucfcrime} and
XD-Violence~\cite{xdviolence} are standard weakly-supervised benchmarks,
with the latter providing aligned audio for every clip.
Recent methods such as MACIL-SD~\cite{macilsd} and UR-DMU~\cite{urdmu}
exploit audio-visual cues with multiple-instance learning and dual memory
banks to localize abnormal snippets.
These works treat audio-visual fusion as the \emph{end task}: produce a
binary anomaly score per snippet.
We reuse the same intuition---that motion and audio energy spike around
salient events---but use these signals as a \emph{front-end} for evidence
selection that is decoupled from any particular downstream task and from
the anomaly label space.

\subsection{Summary}

Prior work either (i) consumes uniformly sampled frames with
ever-larger VLMs, (ii) trains query-aware or agent-based localizers,
or (iii) fuses audio and vision to directly produce anomaly labels.
Recent selectors increasingly avoid training and repeated VLM calls,
but remain query-conditioned and vision-centric. EARS instead combines
\emph{training-free}, \emph{single-pass}, \emph{query-free}, and
\emph{audio-visual} selection. The evidence score is additive with
semantic/query-aware signals (CLIP/SigLIP, learned rankers) and
serves as a cheap default when those signals are unavailable.

\section{Method}
\label{sec:method}

\subsection{Problem Setup}
\label{sec:problem}

Let an input video be a sequence of $T$ RGB frames
$\mathcal{V} = \{f_t\}_{t=1}^{T}$ with an accompanying mono audio waveform
$a \in \mathbb{R}^{L}$ sampled at $f_a$~Hz.
Given a frame budget $K \ll T$ imposed by the downstream
vision-language model (VLM), our goal is to select a frame subset
$\mathcal{F} \subset \mathcal{V}$ with $|\mathcal{F}| = K$ that maximizes
the VLM's downstream performance.

Formally, denoting the downstream evaluation score (e.g.\ AP, accuracy)
by $\Phi$ and the VLM by $g_\theta$, the ideal selector solves
$\mathcal{F}^{\star} = \arg\max_{\mathcal{F} \subset \mathcal{V},\, |\mathcal{F}|=K}\Phi(g_\theta(\mathcal{F}))$,
which is intractable. Instead, we
approximate it by selecting frames that maximize a cheap proxy
$A_t \in [0,1]$ of \emph{multimodal information density}, computed from
modality-intrinsic signals that require no training and no VLM
forward pass.

Our pipeline (Fig.~\ref{fig:ic}) has three stages:
(1) extract a motion signal $M_t$ and an audio signal $E_t$;
(2) fuse them into an evidence score $A_t$;
(3) select $K$ frames using an Information Compression (IC) operator
that enforces both high evidence and temporal diversity.

\subsection{Multimodal Signal Extraction}

\paragraph{Motion signal.}
Dense optical flow is computed between consecutive RGB frames using
the classical Farneback algorithm~\cite{farneback2003} as
implemented in OpenCV's dense optical-flow routine, evaluated on a
strided subset (stride $s\!=\!5$, capped at $2000$ frames per clip)
and cached so that sweeps over $K$ reuse the same signal.
Let $\mathbf{u}_t(p) \in \mathbb{R}^{2}$ denote the flow vector at
spatial location $p$ in frame $t$ and $P$ the set of pixel locations.
The per-frame motion magnitude is
\begin{equation}
M_t \;=\; \frac{1}{|P|} \sum_{p \in P} \lVert \mathbf{u}_t(p) \rVert_{2}.
\label{eq:motion}
\end{equation}
Using flow magnitude rather than raw frame difference reduces sensitivity
to illumination changes, although global camera motion can still produce
spurious evidence.

\paragraph{Audio signal.}
The audio waveform $a$ is split into non-overlapping windows of length
$w = f_a / \mathrm{fps}$ aligned with the video timestamps.
For the window centered at frame $t$, containing $N = w$ samples
$\{a_i\}_{i=1}^{N}$, we compute the short-time energy
\begin{equation}
E_t \;=\; \frac{1}{N} \sum_{i=1}^{N} a_i^{2}.
\label{eq:audio}
\end{equation}
Squared amplitude (rather than absolute value) is used so that $E_t$
correlates with acoustic power, which empirically tracks salient
events such as shouts, impacts and gunshots in our target datasets.
For silent or audio-corrupted clips, $E_t$ is set to zero and the
pipeline degenerates to a motion-only variant (cf.~Sec.~\ref{sec:exp}).

\subsection{Evidence Score}

To make $M_t$ and $E_t$ comparable across videos with different motion
and acoustic profiles, each signal is normalized per-video to
$[0,1]$ via min-max scaling:
\begin{equation}
\begin{aligned}
\tilde{M}_t &= \frac{M_t - \min_{t'} M_{t'}}
                    {\max_{t'} M_{t'} - \min_{t'} M_{t'}}, \\
\tilde{E}_t &= \frac{E_t - \min_{t'} E_{t'}}
                    {\max_{t'} E_{t'} - \min_{t'} E_{t'}}.
\end{aligned}
\label{eq:norm}
\end{equation}
The two normalized signals are then fused into a scalar
\emph{evidence score} by a linear combination,
\begin{equation}
A_t \;=\; \alpha\, \tilde{E}_t \,+\, \beta\, \tilde{M}_t ,
\label{eq:evidence}
\end{equation}
where $\alpha,\beta\!\geq\!0$ balance the two modalities. We use a
\emph{single} configuration everywhere,
$(\alpha,\beta,\tau)\!=\!(1,1,0.5\,\text{s})$, where $\tau$ is the IC gap
of Sec.~\ref{sec:ic}. It is not hand-picked: a grid search on a held-out
split disjoint from every test video selects it independently on both
datasets (Sec.~\ref{sec:exp}). EARS therefore has no per-dataset
hyper-parameters, and the equal weighting is a consequence of that search
rather than an assumption about modality availability in that original study.

\paragraph{Updated controls.}
New controls use up to 256 full-video motion candidates, one-second
mean-absolute-amplitude audio, and sigmoid fusion. Alignment and
strict decoding are specified in the supplementary.

\paragraph{Interpretation.}
$A_t$ acts as a proxy for multimodal information density: motion captures
visual activity and audio captures acoustic activity, so temporal peaks of
$A_t$ indicate that either or both modalities rise above their per-video
baselines.
Unlike event-detection methods that learn a discriminator against a
predefined label space, our score is \emph{label-agnostic} and produced
without any training; it is intentionally a \emph{generic} signal
(motion+audio energy) rather than a query- or task-specific one.

\paragraph{Evidence vs.\ events.}
We deliberately use the term \emph{evidence} rather than \emph{event}:
$A_t$ does not classify what happens, only that something
information-bearing happens.
This distinction matters for downstream reasoning, where the VLM is the
component that has to decide \emph{what} the evidence implies.

\subsection{Evidence Selection with Information Compression}
\label{sec:ic}

\begin{figure}[t]
    \centering
    \includegraphics[width=\linewidth]{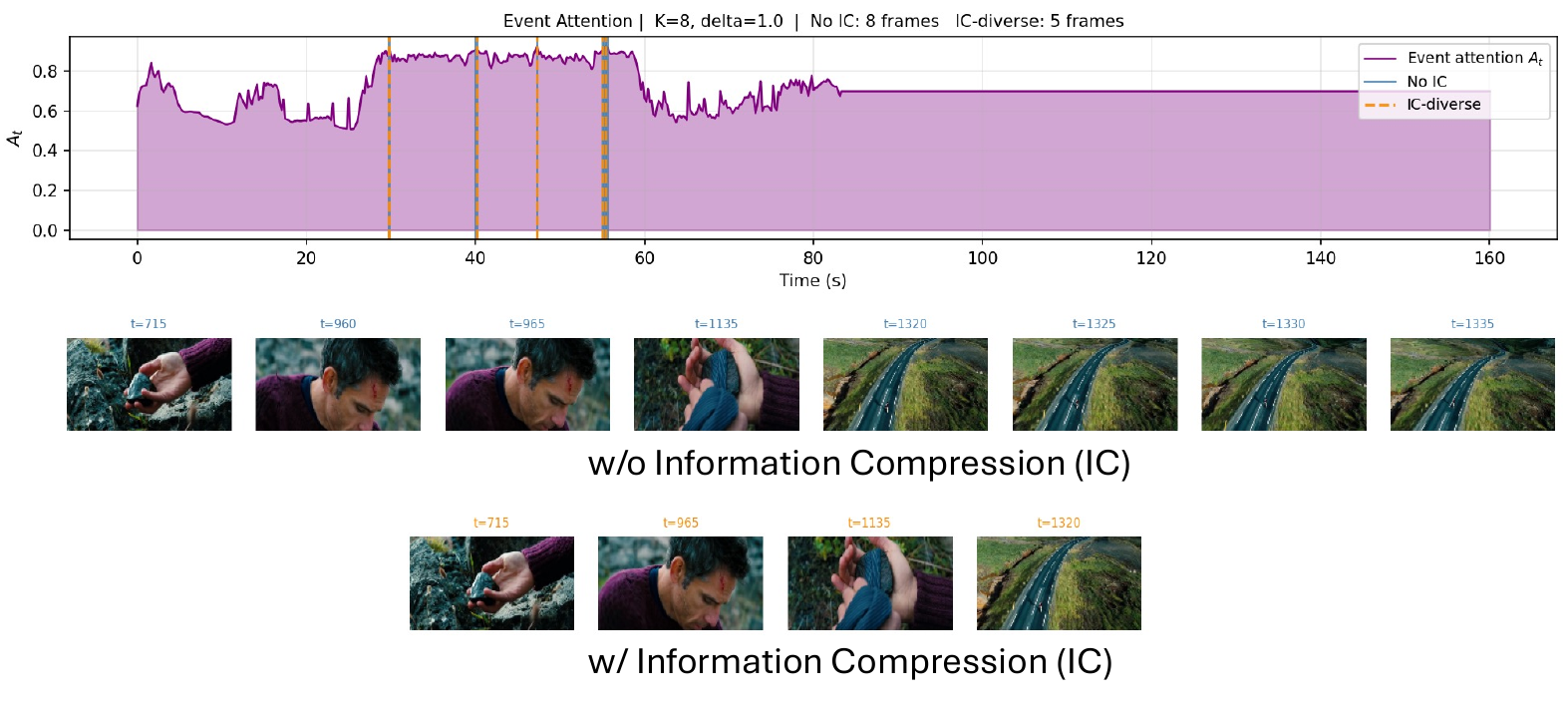}
    \caption{
    \textbf{IC suppresses redundant temporal evidence.} Vanilla top-$K$
    clusters around one peak; a minimum gap $\tau$ retains diverse frames.}
    \label{fig:ic}
\end{figure}

A naive realization of the selection problem is to keep the
$K$ time steps with the highest evidence:
$\mathcal{F}_{\text{top-}K} = \mathrm{TopK}_{t}\, A_t$.
In practice this collapses onto a single high-energy region: adjacent
frames inside one event have nearly identical evidence scores, so the
top-$K$ set is redundant and ignores other informative regions
(top panel of Fig.~\ref{fig:ic}); a sharp onset can trap the whole
budget inside a $<\!2$-second window (supplementary~S1).

We mitigate this with an \textbf{Information Compression (IC)} operator
--- a practical \emph{temporal-NMS / diversity-aware peak-selection}
heuristic, \emph{not} an information-theoretic compression objective
--- that performs greedy non-maximum suppression on the evidence
signal, enforcing a minimum temporal gap $\tau$ between any two
selected frames (the name reflects the conceptual goal of compressing
redundant evidence; a reader who prefers may freely substitute
``Temporal Evidence NMS'' or ``Diversity-Aware Evidence Selection''):

Starting from candidates $\mathcal{C}=\{1,\ldots,T\}$, IC repeatedly
selects $t^\star=\arg\max_{t\in\mathcal{C}}A_t$ and suppresses
$\{t:|t-t^\star|\leq\tau\}$. Algorithm~\ref{alg:ic} gives the complete
procedure, including early termination when no candidates remain.

\begin{algorithm}[t]
\caption{Evidence Selection with Information Compression}
\label{alg:ic}
\small
\begin{algorithmic}[1]
\Require Evidence scores $\{A_t\}_{t=1}^{T}$, budget $K$, gap $\tau$
\State $\mathcal{F} \gets \emptyset$;\quad $\mathcal{C} \gets \{1,\dots,T\}$
\While{$|\mathcal{F}| < K$ \textbf{and} $\mathcal{C} \neq \emptyset$}
    \State $t^{\star} \gets \arg\max_{t \in \mathcal{C}} A_t$
    \State $\mathcal{F} \gets \mathcal{F} \cup \{t^{\star}\}$
    \State $\mathcal{C} \gets \mathcal{C} \setminus \{t : |t-t^{\star}| \leq \tau\}$
\EndWhile
\State \Return $\{f_t:t\in\mathcal{F}\}$ sorted by $t$
\end{algorithmic}
\end{algorithm}

\noindent
The parameter $\tau$ controls the diversity/coverage trade-off: smaller
$\tau$ behaves like vanilla top-$K$, while larger $\tau$ forces the
selected frames to spread out in time. We use the held-out global
$\tau\!=\!0.5$\,s throughout. Ablations in Sec.~\ref{sec:exp} show that
shorter clips or high-budget regimes may prefer weaker diversity
inhibition.
IC is temporal non-maximum suppression rather than an
information-theoretic objective; its connection to token merging
in ViTs~\cite{tome} is an analogy based on discarding near-duplicates.

The output $\mathcal{F}$ of this procedure is the compact frame
subset passed to the downstream VLM
$g_\theta$.
Fusion and IC add only $\mathcal{O}(T)$ work beyond flow and audio-energy
extraction and are negligible in our timing study; the extracted signals
are cached for reuse across budgets and queries.

\ifdefined\issupplement
\section{Additional Experimental Results}
\else
\section{Experiments}
\label{sec:exp}

We report anomaly classification under varying frame budgets $K$ and
compare against query-aware selectors. The supplementary provides
frame-level localization, ablations, backbone transfer, efficiency,
and routing analyses.
Earlier tables and figures retain their original protocols;
Sec.~\ref{sec:exp_updated_controls} reports the updated study separately.

\subsection{Datasets, Tasks and Implementation}
\label{sec:exp_setup}

\paragraph{Datasets.}
We evaluate on two long-form surveillance benchmarks with sparse anomalies:
\textbf{XD-Violence}~\cite{xdviolence} (a local 504-video test subset,
binary violence label) and \textbf{UCF-Crime}~\cite{ucfcrime}
(290 unique test videos; the original tables counted 440 entries,
including repeated normal videos). The updated study deduplicates
and derives binary labels from annotations. We additionally repurpose the
\emph{frame-level} XD-Violence annotations for the localization study
in the supplementary.

\paragraph{Tasks and metrics.}
For \emph{video-level anomaly classification} we report
Accuracy, Precision, Recall, and F1 with the anomaly class as positive.
For \emph{frame-level localization} we report Average Precision (AP) and
AUC against frame-level binary anomaly masks.

\paragraph{VLM and prompts.}
Within each benchmark all variants share the same backbone, prompt,
greedy decoding ($T\!=\!0$), and frame pipeline; \emph{only the
selection strategy differs}. We use Qwen2-VL-7B-Instruct~\cite{qwen2vl}
for UCF-Crime and XD-Violence (binary anomaly prompt in
Table~\ref{tab:main_ucf_xd}; matched multi-choice prompt for the
supplementary ablations and Random sanity check) and
Qwen2.5-VL-7B-Instruct~\cite{qwen25vl} for the
EgoSchema MCQ probe (its MCQ-tuned chat template is required there).
The original studies use the single frozen configuration
$(\alpha,\beta,\tau)\!=\!(1,1,0.5\,\text{s})$ selected on a disjoint
held-out split (supplementary); no hyper-parameter is tuned
per dataset or per budget. New controls fix the same settings and
one binary prompt per cohort, without a new held-out search.

\paragraph{Method-variant glossary.}
Unless stated otherwise, \textbf{EARS} = $A_t\!=\!\alpha\tilde{E}_t\!+\!\beta\tilde{M}_t$ with the frozen global $(\alpha,\beta,\tau)\!=\!(1,1,0.5\,\text{s})$. \textbf{Uniform} is evenly spaced sampling; other variants are defined in the supplementary.

\subsection{Anomaly Classification under Varying Frame Budgets}
\label{sec:exp_main}

Table~\ref{tab:main_ucf_xd} reports both benchmarks under
$K \in \{1,2,4,6,8,10\}$ frames per video, and
Fig.~\ref{fig:f1_vs_k} visualizes the F1 trend.

\begin{table}[t]
\centering
\caption{Original-protocol classification: Qwen2-VL-7B~\cite{qwen2vl},
binary prompt, greedy decoding. UCF includes duplicate entries.
All rows are retained, not directly comparable to updated
Sec.~\ref{sec:exp_updated_controls}. Best per cell is \textbf{bold};
older multi-choice ablations use a separate protocol.}
\label{tab:main_ucf_xd}
\footnotesize
\setlength{\tabcolsep}{3pt}
\resizebox{\columnwidth}{!}{%
\begin{tabular}{l|c|cccc|cccc}
\hline
& & \multicolumn{4}{c|}{\textbf{UCF-Crime} (440 entries)} & \multicolumn{4}{c}{\textbf{XD subset} (504 videos)} \\
Method & $K$ & Acc & P & R & F1 & Acc & P & R & F1 \\
\hline
Uniform     & 1  & \textbf{0.398} & \textbf{1.000} & \textbf{0.086} & \textbf{0.159} & 0.665 & 0.927 & 0.186 & 0.310 \\
\textbf{EARS} & 1 & 0.382 & 0.909 & 0.069 & 0.128 & \textbf{0.819} & \textbf{0.952} & \textbf{0.583} & \textbf{0.723} \\
\hline
Uniform     & 2  & \textbf{0.459} & \textbf{0.919} & \textbf{0.197} & \textbf{0.324} & 0.730 & 0.886 & 0.382 & 0.534 \\
\textbf{EARS} & 2 & 0.432 & 0.870 & 0.162 & 0.273 & \textbf{0.891} & \textbf{0.946} & \textbf{0.775} & \textbf{0.852} \\
\hline
Uniform     & 4  & \textbf{0.527} & \textbf{0.918} & \textbf{0.310} & \textbf{0.464} & 0.897 & 0.913 & 0.824 & 0.866 \\
\textbf{EARS} & 4 & 0.502 & 0.908 & 0.272 & 0.419 & \textbf{0.931} & \textbf{0.933} & \textbf{0.892} & \textbf{0.912} \\
\hline
Uniform     & 6  & \textbf{0.541} & 0.940 & \textbf{0.324} & \textbf{0.482} & 0.944 & \textbf{0.949} & 0.912 & 0.930 \\
\textbf{EARS} & 6 & 0.516 & \textbf{0.964} & 0.276 & 0.429 & \textbf{0.950} & 0.945 & \textbf{0.931} & \textbf{0.938} \\
\hline
Uniform     & 8  & \textbf{0.543} & 0.932 & \textbf{0.331} & \textbf{0.489} & 0.946 & 0.936 & 0.931 & 0.934 \\
\textbf{EARS} & 8 & 0.523 & \textbf{0.955} & 0.290 & 0.444 & \textbf{0.952} & \textbf{0.937} & \textbf{0.946} & \textbf{0.941} \\
\hline
Uniform     & 10 & \textbf{0.550} & 0.942 & \textbf{0.338} & \textbf{0.497} & \textbf{0.968} & \textbf{0.952} & \textbf{0.971} & \textbf{0.961} \\
\textbf{EARS} & 10& 0.539 & \textbf{0.958} & 0.314 & 0.473 & 0.956 & 0.929 & 0.966 & 0.947 \\
\hline
\end{tabular}
}
\end{table}

\begin{figure}[t]
\centering
\includegraphics[width=\linewidth]{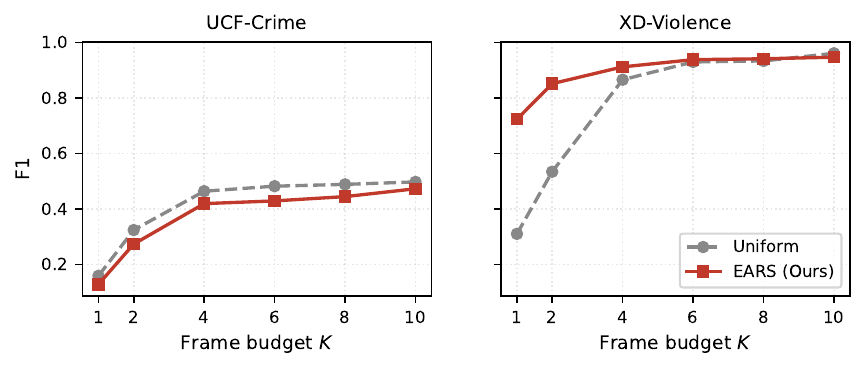}
\vspace{-5pt}
\caption{F1 vs frame budget $K$ on UCF-Crime (left) and XD-Violence (right),
plotted from the EARS and Uniform columns of Table~\ref{tab:main_ucf_xd}
(original Qwen2-VL-7B protocol, not the updated centered-Uniform study). On
XD-Violence the EARS curve sits above Uniform through $K\!=\!8$
(F1 $+0.41$ at $K\!=\!1$, $+0.32$ at $K\!=\!2$), crossing only at
$K\!=\!10$. On mostly-silent UCF-Crime, Uniform remains modestly
stronger at every budget. This contrast --- durable gains when
audio--motion evidence aligns with the event, negative transfer when it
does not --- is the empirical
signature of EARS being an evidence router rather than a universal
selector.}
\label{fig:f1_vs_k}
\end{figure}

\paragraph{Audio--motion alignment determines the gain.}
On XD-Violence, where sparse events often produce both visible motion
and sound, the small-$K$
regime is exactly the practical operating point a deployed VLM faces.
With centered Uniform in the updated study, F1 is
$0.614 \!\rightarrow\! 0.660$ at $K\!=\!1$ and
$0.791 \!\rightarrow\! 0.817$ at $K\!=\!2$.
All six EARS-minus-Uniform paired intervals include zero; the original
curves do not establish the size of a centered-Uniform gain.

\paragraph{Negative transfer on UCF-Crime.}
On mostly-silent UCF-Crime, EARS trails Uniform by $0.025$--$0.051$ F1
across budgets. Increasing $K$ improves both methods, but subtle anomalies
such as Shoplifting and Stealing often produce no audio or motion spike for
EARS to rank. We retain this negative result because it isolates the
applicability condition: sparse events alone are insufficient; their
evidence must align with the cheap modalities.

\paragraph{High-budget crossover.}
On XD-Violence EARS remains ahead through $K\!=\!8$ and crosses Uniform
only at $K\!=\!10$ ($\text{F1}\!=\!0.947$ vs $0.961$) in the
original protocol. The updated full-video cohort has median duration
99.52\,s (UCF: 59.62\,s), so a sub-30\,s XD median cannot explain this crossover.
As $K\!\to\!N$, all selectors become equivalent because every candidate
frame is retained. We exploit this in
the supplementary with a \emph{selectivity-annealed hybrid} that lets EARS soften its
selectivity as $K$ grows, recovering uniform behavior in the high-budget
limit without any training.

\subsection{Qualitative Analysis}
\label{sec:exp_qualitative}

Fig.~\ref{fig:qual_success} shows two XD-Violence wins. In the
\textit{Shooting} clip (a, 50\,s), the violent event is a brief
moment around $t\!\approx\!37$\,s in which a handgun is raised at
the victim. Uniform devotes most of its six-frame budget to
motion-blurred chase shots in which neither the weapon nor the
victim is identifiable; driven by the motion+audio peak, EARS
instead returns four crisp, close-range frames of the gun pointed
at the victim's head---the literal forensic evidence the VLM needs.
In the \textit{Fighting} clip (b, $\sim$3\,min hockey broadcast),
uniform samples generic faceoff and crowd shots; EARS concentrates
on a $\sim$30\,s window covering both an on-ice brawl and a hard
boards body-check. These illustrate where the small-$K$ gap of
Sec.~\ref{sec:exp_main} comes from in practice.

\begin{figure*}[t]
\centering
\begin{minipage}[t]{0.49\textwidth}
\centering
\includegraphics[width=\linewidth]{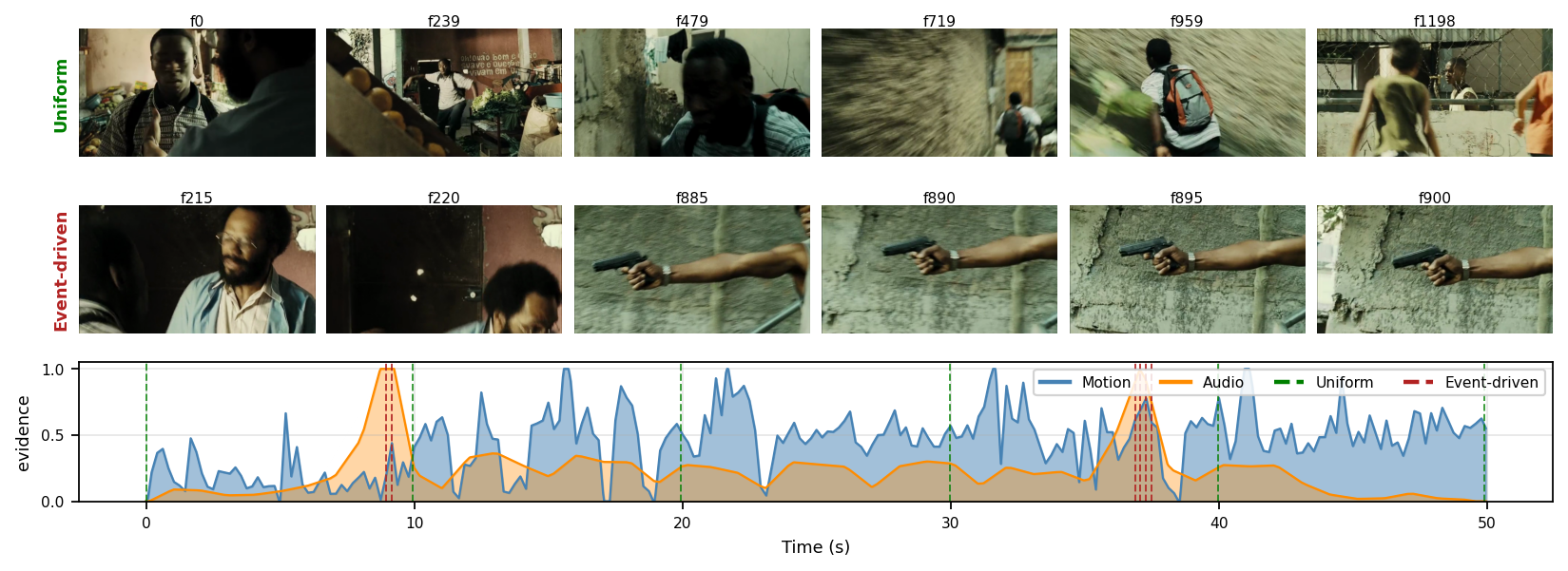}\\[-2pt]
{\footnotesize (a) \textit{Shooting} (50\,s, $K\!=\!6$).}
\end{minipage}\hfill
\begin{minipage}[t]{0.49\textwidth}
\centering
\includegraphics[width=\linewidth]{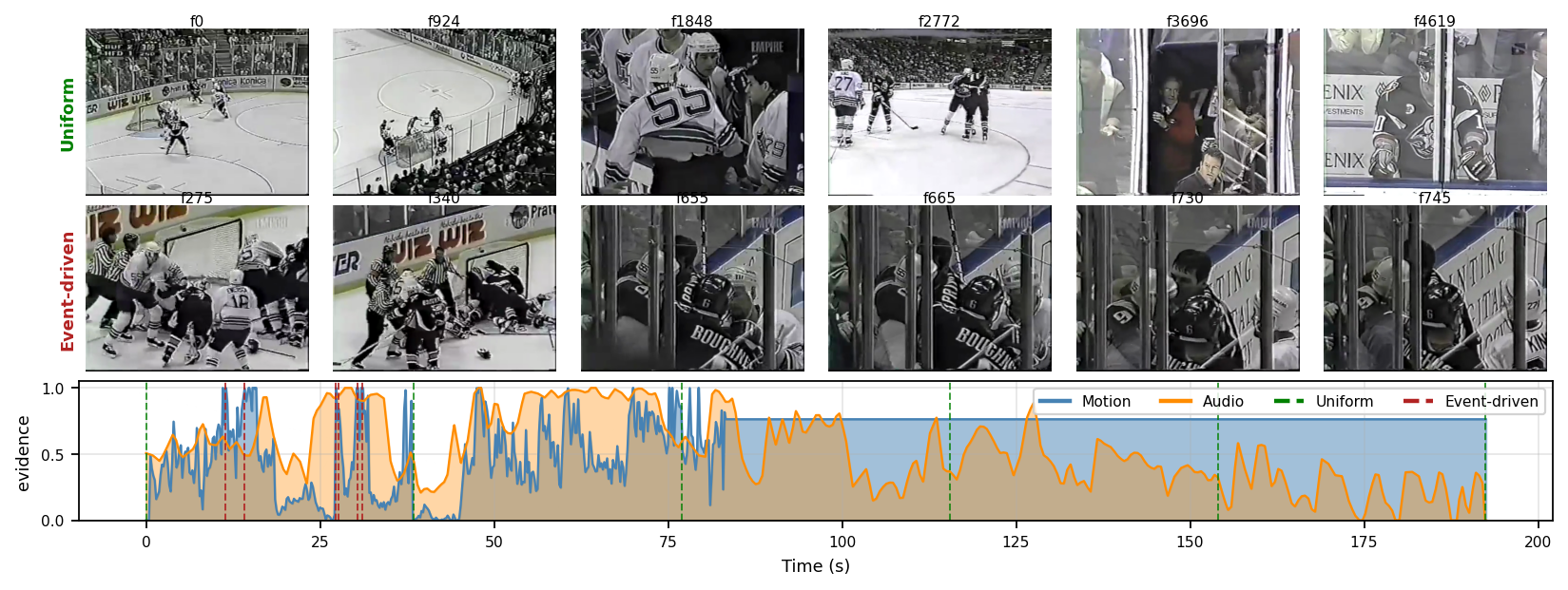}\\[-2pt]
{\footnotesize (b) \textit{Fighting} ($\sim$3\,min, $K\!=\!6$).}
\end{minipage}
\vspace{4pt}
\caption{\textbf{Successful cases.} Uniform frames (top), EARS frames
(middle), and evidence with selection times (bottom). EARS captures the
raised handgun and on-ice brawl that Uniform misses; characteristic
failure modes are shown in the supplementary.}
\label{fig:qual_success}
\end{figure*}

\paragraph{Failure modes.}
Fig.~\ref{fig:qual_failures} exposes two limits of a shallow,
modality-intrinsic prior: one peak can consume the budget, while motion
unrelated to the event can hijack it. IC addresses the former; the latter
requires semantic evidence. Further examples and per-class analysis are
provided in the supplementary.

\begin{figure*}[t]
\centering
\begin{minipage}[t]{0.49\textwidth}
\centering
\includegraphics[width=\linewidth]{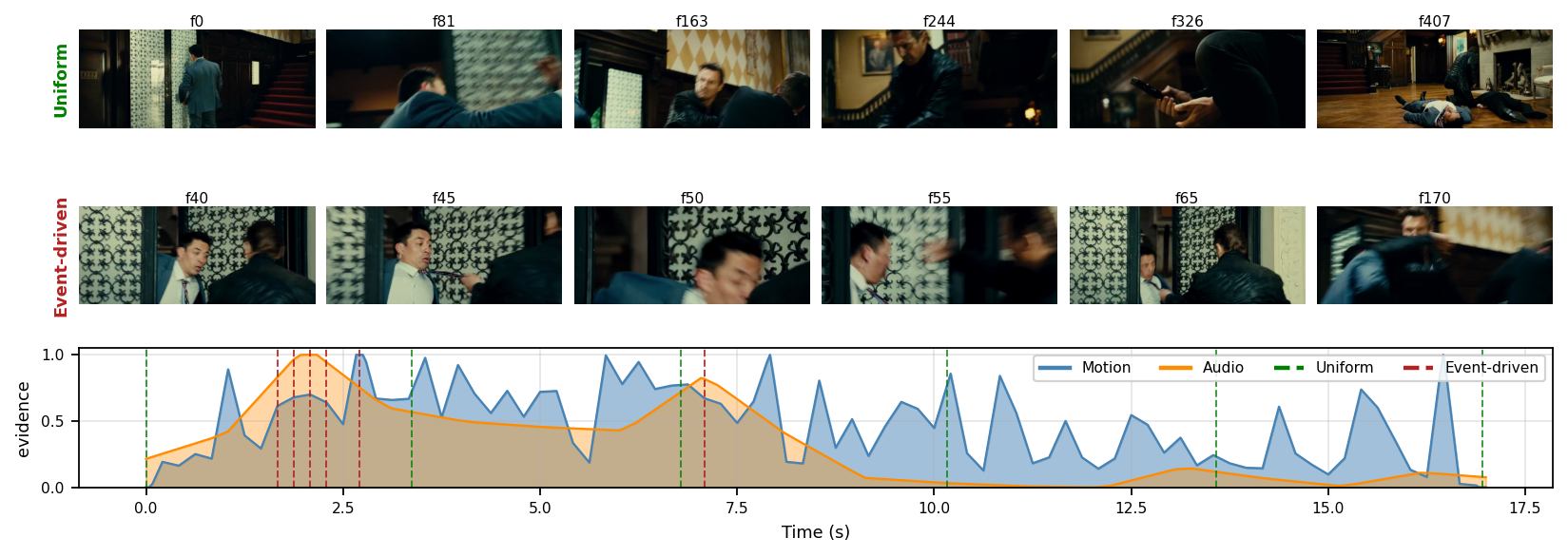}\\[-2pt]
{\footnotesize (a) Peak clustering, XD-Violence \emph{Fighting}.}
\end{minipage}\hfill
\begin{minipage}[t]{0.49\textwidth}
\centering
\includegraphics[width=\linewidth]{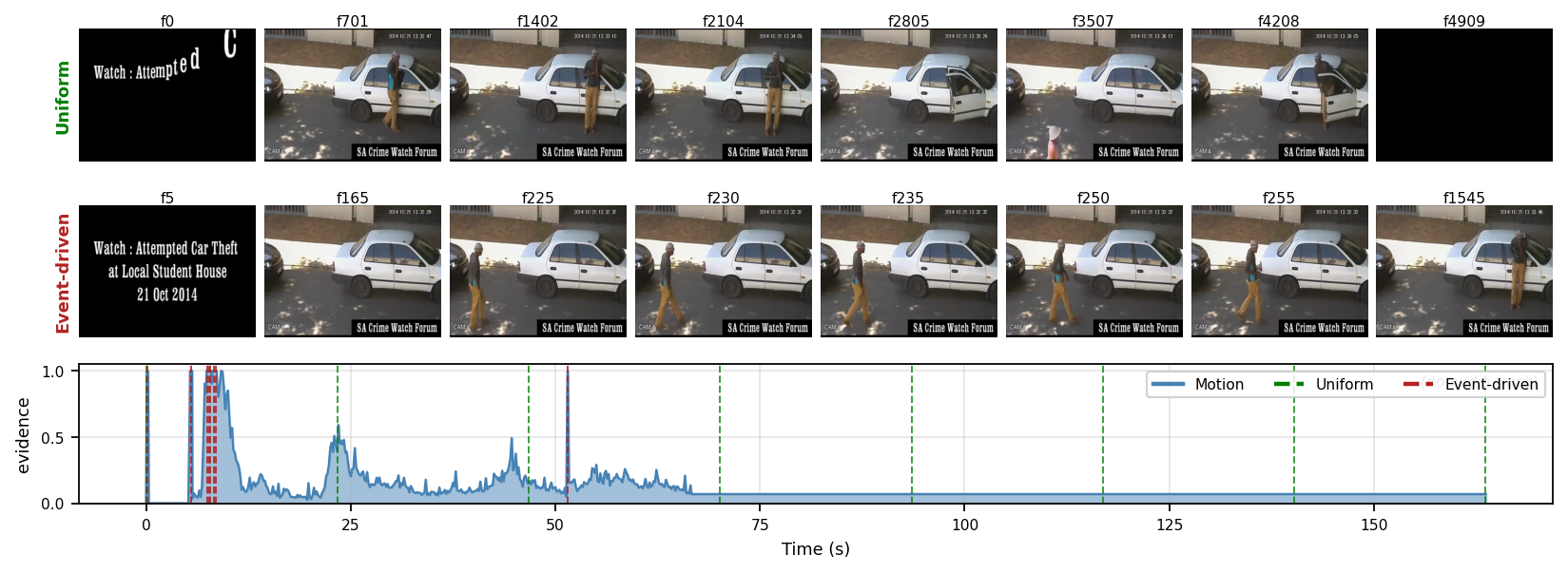}\\[-2pt]
{\footnotesize (b) Proxy hijack, UCF-Crime \emph{Stealing}.}
\end{minipage}
\vspace{2pt}
\caption{\textbf{Characteristic failures.} Uniform (top), EARS (middle),
and evidence (bottom). (a) An early peak clusters EARS picks and misses a
later event. (b) In a silent clip, title-card motion distracts EARS from
the subtle theft.}
\label{fig:qual_failures}
\end{figure*}

\subsection{Comparison with Prior Frame-Selection Methods}
\label{sec:exp_sota}

To contextualize EARS we compare with five recent representative
video-LLM frame selectors: AKS~\cite{aks}, SeViLA~\cite{sevila},
VideoAgent~\cite{videoagent}, M-LLM-FS~\cite{mllmfs}, and
VideoTree~\cite{videotree}. Among these, AKS~\cite{aks} and a
SigLIP top-$K$ selector~\cite{siglip} are the only two that can be
\emph{faithfully} reproduced on our query-free binary-anomaly
benchmark without retraining or proprietary components; the
supplementary explains why the remaining methods are not directly
comparable without changing their original assumptions: they are trained
for question-conditioned QA or require an LLM/API in the loop. We
therefore run \emph{both} as query-aware baselines at every budget on
both datasets under our exact backbone, prompt, and split
(Table~\ref{tab:baselines}); all three selectors share the identical
VLM, prompt, frame budgets, and test split, and both query-aware
baselines reuse the same frozen SigLIP relevance as their only scoring
signal --- only the frame-selection rule differs.

Two observations follow. (a)~On the visually-salient, audio-rich
XD-Violence, AKS and SigLIP top-$K$ are given a task-specific
anomaly query (``fighting/shooting/\ldots'') and edge out EARS at every
budget below four frames ($K\!=\!1$: $0.86$ vs.\ $0.72$); EARS instead targets
\emph{query-free} deployment, where such a query is unavailable.
(b)~On the harder, mostly-silent UCF-Crime, the query-aware selectors
lead at $K\!\le\!2$, Uniform leads at $K\!=\!4$, and EARS trails both.
The query-aware selectors behave almost identically to each other and
EARS catches them on XD by $K\!=\!4$. This is the
scope our paper claims: EARS is \emph{not} a universal selector but a
training-free, \emph{query-free} front-end for sparse-event, low-budget
regimes, explicitly \emph{complementary} to query-aware selectors and
combinable with them, for example as a cheap pre-filter.

\begin{table}[t]
\centering\small
\caption{\textbf{Query-aware selector baselines} (F1 at tight budgets;
same Qwen2-VL-7B~\cite{qwen2vl} backbone, prompt,
split, and frozen SigLIP relevance for all rows). The two query-aware
selectors (AKS, SigLIP top-$K$) are strong when a task-specific
anomaly query is available; EARS is the encoder-free, query-free option.
Best per (dataset,$K$) in \textbf{bold}.}
\label{tab:baselines}
\setlength{\tabcolsep}{3pt}
\resizebox{\columnwidth}{!}{%
\begin{tabular}{lcccccc}
\toprule
 & \multicolumn{3}{c}{XD-Violence} & \multicolumn{3}{c}{UCF-Crime} \\
\cmidrule(lr){2-4}\cmidrule(lr){5-7}
Method & $K{=}1$ & $K{=}2$ & $K{=}4$ & $K{=}1$ & $K{=}2$ & $K{=}4$ \\
\midrule
Uniform                       & 0.310 & 0.534 & 0.866 & 0.159 & 0.324 & \textbf{0.464} \\
AKS~\cite{aks} \emph{(q-aware)}   & \textbf{0.861} & \textbf{0.895} & \textbf{0.890} & 0.242 & 0.346 & 0.412 \\
SigLIP top-$K$~\cite{siglip} \emph{(q-aware)} & \textbf{0.861} & 0.889 & \textbf{0.890} & 0.242 & 0.297 & 0.359 \\
\textbf{EARS} \emph{(q-free)}     & 0.723 & 0.852 & \textbf{0.912} & 0.128 & 0.273 & 0.419 \\
\bottomrule
\end{tabular}
}
\end{table}

\fi
\ifdefined\issupplement

\begin{table}[H]
\centering\footnotesize
\caption{Selector requirements and matched evaluations. \cmark\ marks
the two query-aware selectors evaluated under EARS's exact backbone,
prompt, and split (Table~\ref{tab:baselines}). Methods designed for
question-conditioned long-video QA require a task query and, in some
cases, trained or proprietary components; the concurrent 2026 methods
were published after our evaluation was completed.}
\label{tab:feas}
\setlength{\tabcolsep}{3pt}
\begin{tabular}{lccc}
\toprule
Selector & Per-task & Extra model / API & Repro. \\
(Sec.~\ref{sec:related}) & query & in the loop & here \\
\midrule
AKS~\cite{aks}          & text     & SigLIP (frozen)              & \cmark \\
SigLIP top-$K$~\cite{siglip} & text & SigLIP (frozen)             & \cmark \\
M-LLM-FS~\cite{mllmfs}  & implicit & VLM per frame                & partial \\
SeViLA~\cite{sevila}    & QA       & BLIP-2 loc.\ (\emph{trained}) & \xmark \\
VideoTree~\cite{videotree} & QA    & LLM-guided CLIP tree         & \xmark \\
VideoAgent~\cite{videoagent} & QA  & GPT-4 agent (API)            & \xmark \\
WFS-SB~\cite{wfssb2026}   & text  & CLIP/DINO features           & \xmark \\
EFS~\cite{efs2026}        & text  & DINO + CLIP features         & \xmark \\
MEC~\cite{mec2026}        & text  & sparse visual index          & \xmark \\
\midrule
\textbf{EARS (ours)}    & \emph{none} & \emph{none}               & \cmark \\
\bottomrule
\end{tabular}
\end{table}

\paragraph{Random baseline and seed variance.}
Because Uniform and EARS are deterministic (evenly-spaced sampling;
greedy $T\!=\!0$ selection), only a random baseline carries seed
variance. We therefore run a same-backbone Random-7B baseline under the
\emph{identical} binary prompt, backbone, and split as
Table~\ref{tab:main_ucf_xd}, repeated over five seeds, so its numbers
are directly comparable (Table~\ref{tab:random_seeds}). Two
observations follow. (a)~The per-seed std is small ($\le\!0.027$ F1),
whereas EARS exceeds Random by $0.05$--$0.10$ F1 across XD
$K\!\le\!4$; the gain is not a lucky draw. (b)~On UCF, both Random and
EARS trail Uniform, reinforcing the negative-transfer result when the
cheap modalities do not align with the event.

\begin{table}[H]
\centering\footnotesize
\caption{\textbf{Random-7B seed variance} (five seeds; same Qwen2-VL-7B
backbone, binary prompt, and split as Table~\ref{tab:main_ucf_xd};
mean$\pm$std F1). Uniform and EARS are deterministic and repeated from
Table~\ref{tab:main_ucf_xd} for reference. Best per
(dataset,$K$) in \textbf{bold}.}
\label{tab:random_seeds}
\setlength{\tabcolsep}{3pt}
\resizebox{\columnwidth}{!}{%
\begin{tabular}{lcccccc}
\toprule
 & \multicolumn{3}{c}{XD-Violence} & \multicolumn{3}{c}{UCF-Crime} \\
\cmidrule(lr){2-4}\cmidrule(lr){5-7}
Method & $K{=}1$ & $K{=}2$ & $K{=}4$ & $K{=}1$ & $K{=}2$ & $K{=}4$ \\
\midrule
Uniform \emph{(det.)} & 0.310 & 0.534 & 0.866 & \textbf{0.159} & \textbf{0.324} & \textbf{0.464} \\
Random & $0.627_{\pm.024}$ & $0.767_{\pm.012}$ & $0.863_{\pm.007}$ & $0.148_{\pm.027}$ & $0.320_{\pm.024}$ & $0.414_{\pm.013}$ \\
\textbf{EARS} \emph{(det.)} & \textbf{0.723} & \textbf{0.852} & \textbf{0.912} & 0.128 & 0.273 & 0.419 \\
\bottomrule
\end{tabular}
}
\end{table}

Table~\ref{tab:sota_qual}
then summarizes the design properties of representative selectors,
and a long-video QA probe on EgoSchema~\cite{egoschema}
(Table~\ref{tab:egoschema}) confirms that the sparse-event prior is,
by design, \emph{not} a universal frame router for dense-narrative QA;
EARS should be read as complementary to semantic/query-aware selectors
rather than universally superior.

\begin{table}[t]
\centering
\caption{\textbf{Frame-selection methods for video LLMs.} Training-free
semantic selectors have advanced rapidly, but remain query-conditioned
and vision-centric; EARS is explicitly audio-visual and query-free.
$N$\,=\,candidate frames;
V/A/T denote visual, audio, text.}
\label{tab:sota_qual}
\footnotesize
\setlength{\tabcolsep}{4pt}
\resizebox{\columnwidth}{!}{%
\begin{tabular}{l|c|c|c|c|c}
\hline
Method & Train-free & Selection signal & VLM passes / video & Modality & Year \\
\hline
AKS~\cite{aks}                & \cmark & CLIP--query sim.       & 1                  & V+T     & '25 \\
SeViLA~\cite{sevila}          & \xmark & Localizer--VLM         & $K\!+\!1$          & V       & '23 \\
VideoAgent~\cite{videoagent}  & \cmark & Iterative agent loop   & many               & V+T     & '24 \\
M-LLM-FS~\cite{mllmfs}        & \xmark & VLM scoring of frames  & $\ge\!N/K$         & V+T     & '25 \\
VideoTree~\cite{videotree}    & \cmark & Hierarchical CLIP tree & $\log N$           & V       & '25 \\
WFS-SB~\cite{wfssb2026}       & \cmark & Semantic boundaries    & 1                  & V+T     & '26 \\
EFS~\cite{efs2026}            & \cmark & Event anchors + MMR    & 1                  & V+T     & '26 \\
MEC~\cite{mec2026}            & \cmark & Evidence--context rank & 1                  & V+T     & '26 \\
\hline
\rowcolor{gray!15}
\textbf{EARS (ours)}          & \cmark & Motion+audio energy    & \textbf{1}        & \textbf{V+A} & '26 \\
\hline
\end{tabular}
}
\end{table}

\paragraph{When EARS does not help.}
EARS is grounded in the assumption that salient events are temporally
sparse and visually or acoustically distinguishable. On dense-narrative
benchmarks such as EgoSchema~\cite{egoschema}, where every sub-clip
contributes to a single multi-step plot and audio is uninformative
(the released clips have no audio track), this assumption breaks down:
motion-driven sampling concentrates frames on egocentric head-turns
rather than on semantically relevant moments. Table~\ref{tab:egoschema}
quantifies this on the EgoSchema 500-question subset under the same
Qwen2.5-VL-7B backbone: uniform sampling outperforms EARS by
$5.6$--$9.0$ accuracy points at $K\!=\!4$ and $K\!=\!8$, confirming
that event-driven sampling is the right inductive bias \emph{precisely
when events are sparse}, and is intentionally not a universal
replacement for uniform coverage on dense-narrative QA. We expect a similar pattern on other generalist long-video QA
suites (Video-MME, LongVideoBench, MLVU), where most questions probe
dense narrative or procedural context rather than a single sparse event;
porting EARS to those benchmarks requires the semantic evidence channel
discussed in Sec.~\ref{sec:exp_discussion_hybrid} and is the most
important extension of this work.

\begin{table}[t]
\centering
\caption{EgoSchema~\cite{egoschema} MCQ accuracy on the 500-question subset
(Qwen2.5-VL-7B-Instruct~\cite{qwen25vl}, identical preprocessing). \textbf{Bold marks
the better method per column.} EARS underperforms
uniform sampling on this dense-narrative, audio-silent benchmark,
consistent with its sparse-event design assumption.}
\label{tab:egoschema}
\footnotesize
\setlength{\tabcolsep}{6pt}
\begin{tabular}{l|cc}
\hline
Method & $K\!=\!4$ Acc & $K\!=\!8$ Acc \\
\hline
Uniform                              & \textbf{0.646} & \textbf{0.692} \\
\textbf{EARS (ours)}                & 0.590         & 0.602 \\
\hline
$\Delta$ (EARS $-$ Uniform)         & $-0.056$      & $-0.090$ \\
\hline
\end{tabular}
\end{table}

\subsection{Backbone Generality}
\label{sec:exp_backbone}

A recurring question is whether the small-$K$ EARS gain is an artifact
of one particular VLM. We therefore re-run the \emph{entire} protocol,
unchanged, on three additional backbones spanning two model families
and three generations: Qwen2.5-VL-7B~\cite{qwen25vl},
Qwen3-VL-8B~\cite{qwen3vl}, and the cross-family
LLaVA-NeXT-Video-7B~\cite{llavanextvideo}.
Only the VLM is swapped; the cached motion/audio signals, the evidence
score, and IC are identical. Table~\ref{tab:backbone} reports F1 at the
tight budgets $K\!\in\!\{1,2,4\}$.

\begin{table}[t]
\centering\small
\caption{\textbf{Backbone transfer stress test.} F1 (Uniform vs.\ EARS)
for three additional VLMs under one selector configuration fixed across
this transfer study;
within this table only the VLM changes. The tight-budget gain transfers to
newer Qwen generations and the cross-family
LLaVA at $K\!\le\!4$ on XD. Gains \textbf{bold} where EARS $>$ Uniform.
$^{\dagger}$LLaVA-NeXT-Video-7B collapses to F1~$0$ at $K\!\ge\!8$ on XD
(its frame-count sweet spot is narrow) and is near the floor on
UCF-Crime, which it cannot solve; we therefore read the cross-family
evidence only where the backbone is competent (XD, $K\!\le\!4$).}
\label{tab:backbone}
\setlength{\tabcolsep}{3.5pt}
\resizebox{\columnwidth}{!}{%
\begin{tabular}{llcccccc}
\toprule
 & & \multicolumn{3}{c}{XD-Violence} & \multicolumn{3}{c}{UCF-Crime} \\
\cmidrule(lr){3-5}\cmidrule(lr){6-8}
Backbone & Sel. & $K{=}1$ & $K{=}2$ & $K{=}4$ & $K{=}1$ & $K{=}2$ & $K{=}4$ \\
\midrule
\multirow{2}{*}{Qwen2.5-VL-7B}
 & Unif.  & 0.136 & 0.302 & 0.675 & 0.014 & 0.198 & 0.297 \\
 & EARS   & \textbf{0.561} & \textbf{0.656} & \textbf{0.758} & \textbf{0.129} & 0.181 & 0.193 \\
\midrule
\multirow{2}{*}{Qwen3-VL-8B}
 & Unif.  & 0.433 & 0.599 & 0.873 & 0.151 & 0.325 & 0.418 \\
 & EARS   & \textbf{0.798} & \textbf{0.873} & \textbf{0.906} & \textbf{0.351} & \textbf{0.381} & 0.397 \\
\midrule
\multirow{2}{*}{LLaVA-NeXT-7B$^{\dagger}$}
 & Unif.  & 0.357 & 0.545 & 0.764 & --- & --- & --- \\
 & EARS   & \textbf{0.693} & \textbf{0.788} & \textbf{0.811} & --- & --- & --- \\
\bottomrule
\end{tabular}
}
\end{table}

Together with the frozen Qwen2-VL result in Table~\ref{tab:main_ucf_xd},
the pattern spans four backbones. On XD-Violence, EARS lifts $K\!=\!1$
F1 by $+0.34$--$+0.43$ (Qwen2-VL $0.31\!\to\!0.72$, Qwen2.5-VL
$0.14\!\to\!0.56$, Qwen3-VL $0.43\!\to\!0.80$, LLaVA
$0.36\!\to\!0.69$), confirming that the effect
is a property of \emph{which frames are shown}, not of a specific VLM.
The separate transfer setting improves UCF at $K\!=\!1$ on Qwen2.5/3,
but the frozen global Qwen2 result is negative; UCF is therefore not a
general gain. Two
caveats remain: the newest Qwen3-VL is already more frame-robust, so
the \emph{absolute} margin shrinks as the backbone improves; and the
cross-family LLaVA-NeXT-Video-7B is competent only in a narrow frame
band (F1 collapses to $0$ at $K\!\ge\!8$ on XD) and effectively cannot
solve UCF-Crime, so we restrict its evidence to XD $K\!\le\!4$, where it
still shows the same routing benefit. The routing claim thus holds
wherever the underlying VLM is competent at all.

\subsection{Frame-Level Temporal Localization}
\label{sec:exp_framelevel}

We additionally evaluate EARS on the frame-level annotations of
XD-Violence (Table~\ref{tab:framelevel}; PR/ROC curves are deferred
to the supplementary material). Each row
uses the same downstream VLM but a different selection strategy.

\begin{table}[t]
\centering
\caption{Frame-level localization on XD-Violence. The raw evidence proxy
$A_t$ is a weak \emph{detector} (AP\,$=$\,0.393) but, used as a
\emph{router} for the VLM, it lifts AP by $+6.3$ over uniform sampling
and a further $+1.9$ when combined with Information Compression.}
\label{tab:framelevel}
\footnotesize
\begin{tabular}{l|cc}
\hline
Method & AP $\uparrow$ & AUC $\uparrow$ \\
\hline
$A_t$ proxy alone (no VLM) & 0.393 & 0.583 \\
Uniform + VLM (baseline) & 0.881 & 0.921 \\
EARS, evidence + VLM (no IC) & 0.925 & 0.938 \\
\textbf{EARS, evidence + IC + VLM} & \textbf{0.944} & \textbf{0.951} \\
\hline
\end{tabular}
\end{table}

The $A_t$-only row is intentional: it demonstrates that the gain does
\emph{not} come from $A_t$ being a strong classifier (it is not).
Rather, $A_t$ is a strong \emph{selector}: it tells the VLM \emph{where to
look}, and the VLM tells us \emph{what is there}.

\subsection{Ablations}
\label{sec:exp_ablation}

We ablate four design choices: (i) modality and fusion weights;
(ii) the Information Compression gap $\tau$; (iii) audio robustness; and
(iv) the optical-flow estimator behind the motion channel.
All ablations use UCF-Crime at $K=2$ unless stated otherwise.

\paragraph{Modality and fusion weights.}
We sweep $(\alpha,\beta)$ over five settings
(Table~\ref{tab:ablation_combined}~(left)). The two datasets give sharply
different verdicts. On UCF-Crime, where most test clips have no usable
audio, motion-only wins (F1 $0.411$ vs.\ $0.291$ for audio-only); any
non-zero audio weight is silently ignored because $E_t$ is flat. On
XD-Violence the picture inverts: audio-only ($F_1\!=\!0.654$) beats
motion-only ($0.593$), and balanced fusion is best overall
($F_1\!=\!0.688$). Balanced fusion is therefore the only setting that is
never harmful: it is optimal on XD and within $0.017$ F1 of motion-only on
UCF, where the audio term is inert rather than misleading. This is the
same configuration the held-out search arrives at without seeing any test
video.

\begin{table}[t]
\centering
\caption{Ablations \emph{(matched multi-choice prompt under the same
Qwen2-VL-7B backbone as Table~\ref{tab:main_ucf_xd}; absolute F1
values are not directly comparable to Table~\ref{tab:main_ucf_xd}
because the two tables use different prompts, but the relative
rankings within each ablation are)}. \textbf{Left:} modality fusion weights
$(\alpha,\beta)$ at the per-dataset budget. Motion dominates the
silent-surveillance UCF set; balanced fusion dominates the multimodal
XD set. \textbf{Right:} IC gap $\tau$ at $K\!=\!4$. A small non-zero
$\tau$ is preferred on long, sparse UCF-Crime clips; vanilla top-$K$
($\tau\!=\!0$) is optimal on short XD-Violence clips.}
\label{tab:ablation_combined}
\footnotesize
\setlength{\tabcolsep}{3pt}
\begin{minipage}{0.52\linewidth}\centering
\resizebox{\linewidth}{!}{%
\begin{tabular}{l|cc|cc}
\hline
 & \multicolumn{2}{c|}{UCF $K\!=\!2$} & \multicolumn{2}{c}{XD $K\!=\!6$} \\
$(\alpha,\beta)$ & Acc & F1 & Acc & F1 \\
\hline
$(1,0)$ audio-only   & 0.436 & 0.291          & 0.790 & 0.654 \\
$(0,1)$ motion-only  & \textbf{0.491} & \textbf{0.411} & 0.758 & 0.593 \\
$(1,1)$ balanced     & 0.482 & 0.394          & \textbf{0.802} & \textbf{0.688} \\
$(2,1)$ audio-heavy  & 0.482 & 0.394          & 0.796 & 0.671 \\
$(1,2)$ motion-heavy & 0.484 & 0.398          & 0.786 & 0.652 \\
\hline
\end{tabular}
}
\end{minipage}\hfill
\begin{minipage}{0.46\linewidth}\centering
\resizebox{\linewidth}{!}{%
\begin{tabular}{c|cc|cc}
\hline
 & \multicolumn{2}{c|}{UCF $K\!=\!4$} & \multicolumn{2}{c}{XD $K\!=\!4$} \\
$\tau$ (s) & Acc & F1 & Acc & F1 \\
\hline
0    & 0.468 & 0.350          & \textbf{0.849} & \textbf{0.778} \\
0.5  & \textbf{0.500} & \textbf{0.427} & 0.817 & 0.716 \\
1    & 0.473 & 0.376          & 0.796 & 0.671 \\
2    & 0.495 & 0.413          & 0.770 & 0.611 \\
4    & 0.486 & 0.405          & 0.750 & 0.563 \\
\hline
\end{tabular}
}
\end{minipage}
\end{table}

\paragraph{IC gap $\tau$.}
Sweeping $\tau \in \{0,0.5,1,2,4\}$\,s at $K\!=\!4$
(Table~\ref{tab:ablation_combined}, right) shows a dataset-dependent
optimum: on UCF-Crime a small gap ($\tau\!=\!0.5$\,s) lifts F1 by
$+0.077$ over $\tau\!=\!0$ by suppressing near-duplicate picks around
a single motion spike; on XD-Violence, where clips are short and
evidence is already concentrated, $\tau\!=\!0$ wins. The realized
frame count drops from $4.0$ at $\tau\!=\!0$ to $1.0$--$1.4$ at
$\tau\!=\!4$, foreshadowing the adaptive-$K$ gains of
Sec.~\ref{sec:exp_efficiency}. With $\alpha\!=\!0$ EARS degrades
gracefully to a motion-only sampler when audio is missing, matching
the motion-only row of Table~\ref{tab:ablation_combined}~(left).
The per-class breakdown (Supp.~\S S5, Fig.~S3) confirms the
inductive bias: wins on motion/sound-driven classes (Assault $+0.67$,
Robbery $+0.40$, Fighting, Shooting), losses on subtle or
isolated-flash classes (Arrest $-0.20$, Arson $-0.33$).

\paragraph{Robustness to the optical-flow estimator.}
To verify that the routing gain is a property of the motion
\emph{evidence} rather than of one particular flow algorithm, we
recompute the entire motion channel $M_t$ on XD-Violence with the
variational TV-L1 optical flow~\cite{tvl1} in place of the default
Farneback estimator, holding the audio channel, fusion weights
$(\alpha,\beta)$, IC gap $\tau$, backbone, and prompt fixed --- only the
flow estimator changes. This estimator-only stress test uses the original
fixed selector setting. The resulting EARS F1 stays within $0.024$ of
the Farneback pipeline at every tight budget (XD $K\!=\!1$: $0.708$
vs.\ $0.732$; $K\!=\!2$: $0.827$ vs.\ $0.843$; $K\!=\!4$: $0.885$
vs.\ $0.888$), and both remain far above Uniform
($0.310/0.534/0.866$). The evidence router is thus robust to the flow
estimator: what matters is that motion energy localizes the sparse
event, not which optical-flow method computes it.

\paragraph{Held-out configuration selection (no test-set tuning).}
A selector with per-dataset weights could be encoding test-set knowledge.
To rule this out we select $(\alpha,\beta,\tau)$ on a held-out split that
is \emph{strictly disjoint} from every test video (XD-Violence
\texttt{train/}; UCF-Crime training videos), $300$ videos per dataset. We
grid over $\alpha\!\in\!\{0,1,2\}$, $\beta\!=\!1$,
$\tau\!\in\!\{0,0.5,1\}$\,s on \emph{both} datasets, pick the best mean F1
at $K\!\le\!2$, \emph{freeze} the configuration, and only then touch the
test set. Both searches independently select the same setting,
$(\alpha,\beta,\tau)\!=\!(1,1,0.5\,\text{s})$, at held-out F1 $0.769$ (XD)
and $0.557$ (UCF). EARS thus uses one global configuration and has no
per-dataset hyper-parameters.

The two datasets agree on the setting for different reasons, and the
distinction matters. On XD-Violence the choice of $\alpha$ is a real
effect: the spread between the best $\alpha\!=\!1$ and the worst
$\alpha\!=\!0$ configuration is $0.063$ F1, against a $0.025$ baseline
obtained by varying only $\tau$ within a fixed $\alpha$. On UCF-Crime the
corresponding numbers are $0.011$ against $0.047$, i.e.\ $\alpha$ is
inconsequential --- as expected, since $222$ of its $290$ test videos
carry no audio track at all. Equal weighting is selected on UCF because
nothing distinguishes it there, and selected on XD because audio genuinely
helps; a single default is defensible precisely because the one dataset
with an opinion prefers it.

\subsection{Efficiency Analysis}
\label{sec:exp_efficiency}

We analyze two efficiency dimensions: (a) the per-video overhead of
computing $A_t$ relative to a VLM forward pass, and (b) the
\emph{adaptive-$K$} variant of IC in which selection terminates once no
candidate exceeds an evidence threshold (Alg.~\ref{alg:ic} extended).

\paragraph{Pre-processing cost and amortization via caching.}
Audio energy and Farneback optical-flow signals (stride $s\!=\!5$,
OpenCV implementation) are extracted on a single CPU core and cached
to disk ($\sim\!10$\,MB for the full XD-Violence test split), so
all $K$ and $\tau$ sweeps reuse the same per-video signal at zero
re-extraction cost.
On 24 XD-Violence and UCF-Crime videos (single CPU core) we measure
$0.28$\,s for audio, $2.84$\,s for single-pass decode plus Farneback
motion, and $<\!1$\,ms for fusion+top-$K$, totaling $3.12$\,s/video;
frame decode for the $K$ selected frames adds $0.07$\,s (identical to
uniform). Caching amortizes the pre-processing across all $K$ and $\tau$
sweeps---a one-time CPU cost paid against repeated GPU VLM inference.
Because $A_t$ is query-independent, the same cached signal also serves
every future query on that video, which query-conditioned scores cannot.

\paragraph{Paired end-to-end latency.}
The preceding number isolates an optimized front-end. To include the
downstream model, we additionally time EARS and Uniform on the same 24
XD videos at $K\!=\!4$ with Qwen2-VL-7B. Uniform takes
$1.33\!\pm\!0.53$\,s/video end to end ($1.23$\,s VLM); EARS takes
$6.87\!\pm\!2.95$\,s ($1.39$\,s VLM). The $5.54$\,s gap is therefore
almost entirely CPU preprocessing: audio $0.37$\,s, Farneback motion
$5.02$\,s, fusion+IC $0.005$\,s, and selected-frame decode $0.09$\,s.
EARS is not faster than Uniform on a first query; its advantage is that
this query-independent cost is cached and amortized across later queries.

\paragraph{How the front-end compares to a learned selector.}
Table~\ref{tab:efficiency} places EARS's front-end next to the SigLIP
relevance encoder that AKS and SigLIP top-$K$ require, measured on the same
24 XD-Violence videos (mean $299$ frames). Both front-ends must first
decode a strided subset of frames, and how that decode is implemented
dominates the comparison: requesting frames individually through
\texttt{CAP\_PROP\_POS\_FRAMES} restarts the decoder at the nearest
keyframe every time and costs $4.30$\,s/video, whereas walking the stream
once and keeping the wanted frames costs $0.17$\,s for
\emph{bit-identical} output. With the single-pass decode, EARS costs
$3.12$\,s/video against $6.89$\,s for the SigLIP pass. We report both
numbers because the SigLIP baseline still uses per-frame seeking; after
accounting for decode implementation, the front-ends differ by roughly $2\times$
once each is implemented well, not by an order of magnitude.

The durable advantage is qualitative rather than a wall-clock margin:
EARS needs \emph{no learned encoder} (it is CPU-only, where the SigLIP
forward pass wants a GPU), \emph{no per-task query}, and \emph{no extra
VLM forward pass}. VLM-in-the-loop selectors (SeViLA, M-LLM-FS,
VideoAgent) instead add one-to-many additional VLM passes per video, a
cost that dwarfs either front-end.

\begin{table}[t]
\centering\small
\caption{\textbf{Selection and end-to-end cost} (24 XD-Violence videos,
mean $299$ frames/video). Frame decoding is shared by both front-ends and
dominates unless it is done in a single pass; the SigLIP row still uses
per-frame seeking, so the realistic gap is near $2\times$, not the $2.5\times$
the front-end rows show. Paired end-to-end rows include Qwen2-VL-7B at
$K\!=\!4$ (mean$\pm$std).}
\label{tab:efficiency}
\setlength{\tabcolsep}{5pt}
\resizebox{\columnwidth}{!}{%
\begin{tabular}{lccc}
\toprule
Front-end & Encoder & Query & s/video \\
\midrule
SigLIP relevance~\cite{siglip} (AKS, top-$K$) & \cmark & \cmark & 6.89 \\
\textbf{EARS}, per-frame seeking & \xmark & \xmark & 7.83 \\
\textbf{EARS}, single-pass decode & \xmark & \xmark & \textbf{3.12} \\
\midrule
Uniform + VLM, end to end & \xmark & \xmark & $1.33{\pm}0.53$ \\
\textbf{EARS} + VLM, end to end & \xmark & \xmark & $6.87{\pm}2.95$ \\
\midrule
VLM-in-the-loop (SeViLA~\cite{sevila}, M-LLM-FS~\cite{mllmfs}) & \cmark & \cmark & $+\,\ge\!1$ VLM pass \\
\bottomrule
\end{tabular}
}
\end{table}

\paragraph{Adaptive-$K$ Pareto.}
Fig.~\ref{fig:adaptive_k} plots F1 vs.\ average frames-per-video as the
evidence threshold $\theta$ is swept. On UCF-Crime the best operating
point is $\theta\!=\!0.65$, yielding $F_1\!=\!0.419$ at only $1.67$
frames/video---a $58\%$ frame reduction over fixed $K\!=\!4$ with no
F1 loss relative to the best fixed setting ($0.427$ at $\tau\!=\!0.5$,
$3.80$ frames/video). On XD-Violence F1 is nearly flat ($\approx\!0.67$)
over $\theta \in [0.5,0.75]$ while the realized budget shrinks from
$2.42$ to $2.17$ frames/video, indicating that $A_t$ is sharply peaked
on XD and adaptive stopping is essentially free. The IC threshold is
thus a no-retraining knob for frugal-frame deployment.

\begin{figure}[t]
\centering
\includegraphics[width=0.85\linewidth]{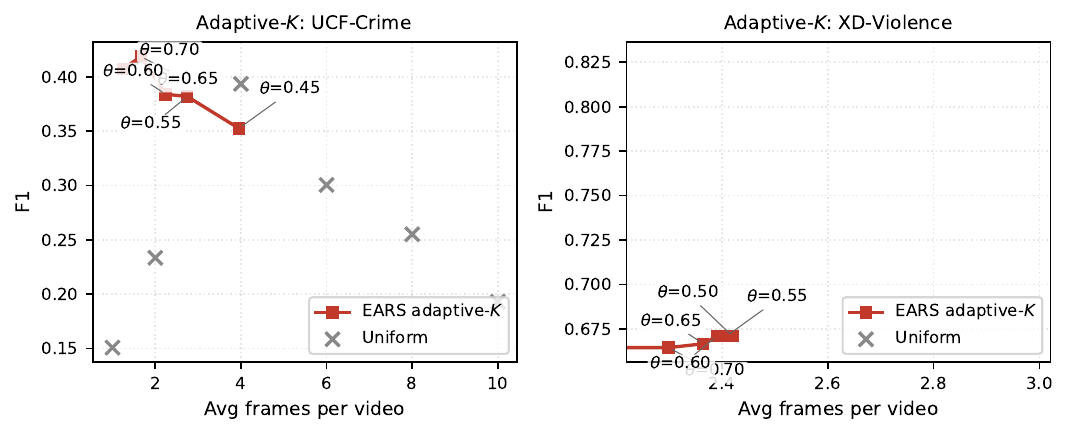}
\vspace{-6pt}
\caption{Adaptive-$K$ Pareto on UCF-Crime (left) and XD-Violence (right).
Sweeping $\theta$ in Alg.~\ref{alg:ic} trades average frames-per-video
against F1; crosses mark fixed-$K$ Uniform at $K\!\in\!\{1,2,4,6,8,10\}$.
UCF matches its best fixed-$\tau$ at $<\!1/2$ the frame cost; XD is flat.}
\label{fig:adaptive_k}
\end{figure}

\subsection{Can Selection Be Routed Per Video?}
\label{sec:exp_routing}

EARS does not help uniformly: it wins decisively at tight budgets on
XD-Violence and loses to Uniform at $K\!\ge\!4$ on UCF-Crime. The natural
response is a router that inspects each video and chooses between
evidence-driven and uniform sampling. Because our signals are
query-independent and already computed, such a router would be
free---which makes it worth asking whether it works. We report a negative
answer, together with the measurement that makes it interpretable.

\paragraph{Setup.}
We evaluate on ten (dataset, $K$) cells: UCF-Crime and XD-Violence at
$K\!\in\!\{1,2,4,8\}$ (F1) and the EgoSchema MCQ probe at
$K\!\in\!\{4,8\}$ (accuracy), spanning sparse-event surveillance and dense
egocentric narrative. Each video is described by eleven statistics of its
own $M_t$ and $E_t$---Gini coefficient, top-mass fraction, normalized
entropy, peak ratio, peakiness, audio activity, and audio--motion
agreement---all computable from the cached signals at no additional cost.
A threshold on their weighted sum selects the arm. Thresholds are chosen
leave-one-dataset-out, so no cell is scored with a threshold fitted on
itself. We compare against the per-video \emph{oracle}, which picks the
better arm with hindsight and upper-bounds any router.

\begin{table}[t]
\centering\small
\caption{\textbf{Routing policies}, mean over ten (dataset, $K$) cells
(F1 for UCF/XD, accuracy for EgoSchema). The oracle shows substantial
headroom, but the learned per-video router underperforms both fixed
policies. Thresholds for both adaptive rows are selected
leave-one-dataset-out.}
\label{tab:routing}
\setlength{\tabcolsep}{6pt}
\begin{tabular}{lc}
\toprule
Policy & Mean score \\
\midrule
Per-video router (modality statistics) & 0.635 \\
Always Uniform & 0.647 \\
Budget rule (EARS iff $K\!\le\!\theta$) & 0.676 \\
\textbf{Always EARS} & \textbf{0.685} \\
\midrule
Per-video oracle (upper bound) & 0.789 \\
\bottomrule
\end{tabular}
\end{table}

\paragraph{The headroom is real; the router cannot reach it.}
Table~\ref{tab:routing} shows an oracle margin of $+0.104$ over always
using EARS, so per-video routing is worth $10$ points if it can be done.
It cannot, from these signals. The router scores $0.635$, below the
$0.647$ of never routing at all. The failure is not a tuning artifact but
degeneracy: the leave-one-dataset-out threshold collapses the router onto
a constant arm, and on XD-Violence it collapses onto the wrong one,
forfeiting the $+0.10$ (at $K\!=\!1$) to $+0.34$ (at $K\!=\!2$) F1 that
EARS would have delivered.

\paragraph{No individual statistic is informative.}
Testing each of the eleven features in each cell for its ability to
predict whether EARS beats Uniform on a given video gives $92$
feature--cell tests. Three reach $p\!<\!0.05$ under a permutation test,
against $4.6$ expected by chance alone; none survives Bonferroni
correction. Mean $|\text{AUC}-0.5|$ ranges from $0.026$ to $0.070$ across
features, where a useful router would need roughly $0.20$. The apparent
exceptions reverse sign across budgets: motion top-mass reaches AUC $0.72$
on UCF at $K\!=\!2$ ($n\!=\!40$) but $0.44$ at $K\!=\!8$. As a sanity
check on the pipeline, every audio feature scores exactly $0.50$ on
EgoSchema, whose clips carry no audio stream.

Two structural reasons underlie this. First, the routable population is
small: $79$--$92\%$ of videos receive identical predictions under both
arms, so the two arms mostly agree and the disagreements are few. Second,
sparsity does not separate the regimes even at the dataset level---the
same score distinguishes dense-narrative EgoSchema from sparse-event
XD-Violence at AUC $0.576$.

\paragraph{Routing on the budget instead of the video.}
Since $K$ governs when EARS stops helping, one might route on $K$ alone.
Choosing the crossover on the same cells gives $0.700$, above always-EARS;
but that threshold is fitted on the data it is scored on. Selected
leave-one-dataset-out it yields $0.676$, again below always-EARS. We
therefore report the constant policy as the strongest one we can validate,
and treat the oracle gap as an open problem: the evidence says the
information needed to route is not present in modality-intrinsic
statistics, so a router would have to look at content, at which point it
is no longer free. This bounds a design space that recent
adaptive-selection work has assumed to be open.

\subsection{Discussion}

Three findings emerge across Tables~\ref{tab:main_ucf_xd}--\ref{tab:framelevel}.
\emph{(i)}~Gains are strongest where sparse events align with
motion or audio: \textbf{XD-Violence} $+0.41$ F1 at $K\!=\!1$
($+0.32$ at $K\!=\!2$), while \textbf{UCF-Crime} shows negative
transfer (subtle, silent anomalies like Shoplifting).
\emph{(ii)}~IC --- temporal NMS on the evidence axis --- is necessary
(AP $0.925\!\to\!0.944$ once added).
\emph{(iii)}~The proxy is a poor classifier (AP $0.393$) but an
excellent VLM router: cheap selector, expensive reasoner.

\paragraph{Scope and hybrid fallback.}
\label{sec:exp_discussion_hybrid}
EARS suits tight budgets on motion/audio-correlated videos; under the
frozen global configuration, XD crosses Uniform only at $K\!=\!10$
(Fig.~\ref{fig:f1_vs_k}). A training-free remedy is the
selectivity-annealed hybrid $S_t=(1{-}\lambda)A_t+\lambda U_t$,
$\lambda=\min(1,K/(\rho^{\!*}N_{\text{eff}}))$ (supp.~S3): pure EARS
and uniform are limits, and the adaptive-$K$ Pareto
(Sec.~\ref{sec:exp_efficiency}) is its dual.


\paragraph{Limitations and future work.}
\textbf{(1)~Stealthy events:} motion+audio cannot rank anomalies with
no spike (e.g.\ Shoplifting); the natural generalization is a semantic
third channel $A_t\!=\!\alpha E_t\!+\!\beta M_t\!+\!\gamma C_t$
(CLIP/SigLIP). \textbf{(2)~Scope:} extending to
Video-MME/LongVideoBench/MLVU requires (1).

\else

\subsection{Matched Semantic Fusion}
\label{sec:exp_updated_controls}

On 504 XD/290 unique UCF videos, we average normalized SigLIP~\cite{siglip}
and EARS scores before identical NMS. XD four-frame F1 rises
$0.847\!\to\!0.889$, below Uniform/EARS/Audio ($0.893/0.898/0.907$);
UCF one-frame F1 drops $0.196$. All six contrasts are in
Table~\ref{tab:updated_hybrid}; full controls and 432 paired
cost trials are in the supplementary.

\begin{table}[!ht]
\centering\footnotesize
\caption{Hybrid minus semantic-NMS: $\Delta$F1 [95\% interval].
Exploratory, unadjusted; 10,000 paired video resamples.}
\label{tab:updated_hybrid}
\setlength{\tabcolsep}{3pt}
\begin{tabular}{cll}
\toprule
$K$ & XD subset & UCF-Crime \\
\midrule
1 & $+0.017\ [-0.023,+0.057]$ & $-0.196\ [-0.302,-0.093]$ \\
2 & $+0.022\ [-0.009,+0.054]$ & $-0.028\ [-0.113,+0.058]$ \\
4 & $+0.042\ [+0.017,+0.067]$ & $+0.009\ [-0.048,+0.066]$ \\
\bottomrule
\end{tabular}
\end{table}

\subsection{Additional Validation}
\label{sec:exp_additional}
Full protocols and per-budget results are in the supplementary.
Table~\ref{tab:main_backbone} tests backbone transfer;
Table~\ref{tab:main_framelevel} separates proxy detection from VLM
localization; Tables~\ref{tab:main_components} and~\ref{tab:main_heldout}
report component and held-out sensitivity. The supplementary retains
the original robustness, scope, and efficiency summary. The
original studies are separate from the updated matched controls.

\begin{table}[t]
\centering
\caption{Backbone-transfer stress test on XD-Violence (F1). A selector
configuration fixed across this transfer study improves every tested VLM
at tight budgets under its original protocol, separate from the new controls.}
\label{tab:main_backbone}
\footnotesize
\setlength{\tabcolsep}{4pt}
\begin{tabular}{lcccc}
\toprule
& \multicolumn{2}{c}{$K=1$} & \multicolumn{2}{c}{$K=2$} \\
Backbone & Unif. & EARS & Unif. & EARS \\
\midrule
Qwen2.5-VL-7B~\cite{qwen25vl} & .136 & \textbf{.561} & .302 & \textbf{.656} \\
Qwen3-VL-8B~\cite{qwen3vl} & .433 & \textbf{.798} & .599 & \textbf{.873} \\
LLaVA-NeXT-7B~\cite{llavanextvideo} & .357 & \textbf{.693} & .545 & \textbf{.788} \\
\bottomrule
\end{tabular}
\end{table}

\begin{table}[t]
\centering
\caption{Frame-level localization on XD-Violence. The evidence proxy
is weak alone but effective for routing frames to the VLM.}
\label{tab:main_framelevel}
\footnotesize
\setlength{\tabcolsep}{5pt}
\begin{tabular}{lcc}
\toprule
Method & AP $\uparrow$ & AUC $\uparrow$ \\
\midrule
$A_t$ proxy (no VLM) & 0.393 & 0.583 \\
Uniform + VLM & 0.881 & 0.921 \\
EARS + VLM (no IC) & 0.925 & 0.938 \\
\textbf{EARS + IC + VLM} & \textbf{0.944} & \textbf{0.951} \\
\bottomrule
\end{tabular}
\end{table}

\begin{table}[t]
\centering
\caption{Component ablations (F1; matched multi-choice protocol).
Fusion uses UCF $K\!=\!2$/XD $K\!=\!6$; IC uses $K\!=\!4$ on both.
Balanced fusion is robust across modalities, while the useful temporal
gap depends on clip structure.}
\label{tab:main_components}
\footnotesize
\setlength{\tabcolsep}{5pt}
\begin{tabular}{llcc}
\toprule
Study & Setting & UCF & XD \\
\midrule
\multirow{3}{*}{Fusion}
 & Audio only $(1,0)$ & .291 & .654 \\
 & Motion only $(0,1)$ & \textbf{.411} & .593 \\
 & Balanced $(1,1)$ & .394 & \textbf{.688} \\
\midrule
\multirow{3}{*}{IC gap}
 & $\tau=0$ s & .350 & \textbf{.778} \\
 & $\tau=0.5$ s & \textbf{.427} & .716 \\
 & $\tau=1$ s & .376 & .671 \\
\bottomrule
\end{tabular}
\end{table}

\begin{table}[t]
\centering
\caption{Held-out sensitivity (mean F1 over $K=1,2$; 300 training
videos/dataset). Both datasets independently select the same global
configuration, and neighboring gaps are nearly tied.}
\label{tab:main_heldout}
\footnotesize
\setlength{\tabcolsep}{6pt}
\begin{tabular}{ccc|cc}
\toprule
$\alpha$ & $\beta$ & $\tau$ (s) & XD & UCF \\
\midrule
0 & 1 & 0.5 & .706 & .548 \\
1 & 1 & 0   & .749 & .530 \\
\textbf{1} & \textbf{1} & \textbf{0.5} & \textbf{.769} & \textbf{.557} \\
1 & 1 & 1   & .768 & .555 \\
\bottomrule
\end{tabular}
\end{table}

\begin{table}[t]
\centering
\caption{Random-sampling control (F1; same frozen Qwen2-VL-7B
protocol as Table~\ref{tab:main_ucf_xd}). Random reports mean$\pm$std
over five seeds; EARS and Uniform are deterministic.}
\label{tab:main_random}
\footnotesize
\setlength{\tabcolsep}{3pt}
\resizebox{\columnwidth}{!}{%
\begin{tabular}{lcccccc}
\toprule
 & \multicolumn{3}{c}{XD-Violence} & \multicolumn{3}{c}{UCF-Crime} \\
\cmidrule(lr){2-4}\cmidrule(lr){5-7}
Method & $K{=}1$ & $K{=}2$ & $K{=}4$ & $K{=}1$ & $K{=}2$ & $K{=}4$ \\
\midrule
Uniform & .310 & .534 & .866 & \textbf{.159} & \textbf{.324} & \textbf{.464} \\
Random & $.627_{\pm.024}$ & $.767_{\pm.012}$ & $.863_{\pm.007}$ & $.148_{\pm.027}$ & $.320_{\pm.024}$ & $.414_{\pm.013}$ \\
\textbf{EARS} & \textbf{.723} & \textbf{.852} & \textbf{.912} & .128 & .273 & .419 \\
\bottomrule
\end{tabular}
}
\end{table}

\paragraph{Stochastic and budget controls.}
Table~\ref{tab:main_random} rules out a fortunate temporal draw: on XD,
EARS exceeds five-seed Random by $0.05$--$0.10$ F1 for $K\!\leq\!4$,
larger than Random's maximum $0.024$ standard deviation there. On UCF,
Random and EARS both trail Uniform, independently reinforcing the negative
transfer finding. We also test adaptive stopping by thresholding the IC
evidence in the matched ablation protocol. On UCF, it retains F1 $0.419$
with $1.67$ frames/video, versus $0.427$ with $3.80$ frames for the best
fixed setting, a $56\%$ frame reduction for $0.008$ F1. On XD, F1 remains
approximately $0.67$ while the realized budget falls from $2.42$ to $2.17$
frames/video over thresholds $0.5$--$0.75$. Thus IC supplies a continuous
accuracy--budget control rather than only a fixed top-$K$ rule; the full
Pareto curves are in the supplementary.

\paragraph{Reproducibility and release.}
Uniform and EARS are deterministic: every reported anomaly result uses
greedy decoding on fixed local cohorts (504 XD videos; the original
440 UCF entries represent 290 unique videos). The new matched study
uses the deduplicated labels. Frozen configurations are serialized before
test evaluation, and tables are generated from saved per-video predictions
rather than transcribed from terminal output. Upon publication, we will
release the code, environment specification, frozen configurations,
held-out split lists, and derived predictions needed to reproduce every
table. UCF-Crime and XD-Violence remain governed by their original licenses;
we will provide official-download and preprocessing scripts rather than
redistribute third-party videos.

\subsection{Discussion}
EARS targets tight budgets where sparse events align with motion or
audio; Uniform catches up as $K$ grows and wins on dense-narrative,
audio-silent EgoSchema. A hybrid can anneal toward Uniform at larger
budgets, while stealthy events require a semantic third channel.
Complete ablations, failures, efficiency, and routing analyses appear in
the supplementary.
Together, the stress tests distinguish robustness from universality.
Flow replacement and backbone transfer preserve the tight-budget XD
effect, while the held-out grid removes test-set configuration choice.
Conversely, EgoSchema and the failed per-video router show that cheap
signals cannot decide when semantic evidence is required. Query-aware
selectors should therefore be preferred when a task query is available;
EARS occupies the complementary setting where no query or selector
encoder is available and the frame budget is the binding constraint.

\fi

\section{Conclusion}

\textbf{EARS} is a training-free, query-free, single-pass frame router
based on cheap motion and audio statistics. At $K\!=\!1$ on
the XD subset its updated F1 is $0.660$ versus centered Uniform's $0.614$
(paired interval includes zero); backbone and five-seed results are
separate original studies.
Although the raw proxy is weak (AP $0.393$),
EARS+IC+VLM reaches frame-level AP $0.944$: routing needs relative temporal
evidence, not a calibrated detector.

Negative results bound the claim: EARS trails Uniform on mostly-silent
UCF-Crime, while dense-narrative EgoSchema requires semantic evidence.

For deployment, EARS trades a one-time CPU preprocessing cost for a
query-independent evidence cache that can be reused across budgets and
future questions. This makes it most useful when repeated reasoning is
required over motion- or audio-correlated events; for silent, semantic, or
dense-narrative videos, Uniform or a query-aware selector remains the safer
fallback. The new semantic--activity hybrid tests this modular extension,
with cohort/budget-dependent benefits.

\FloatBarrier
{
  \small
  \bibliographystyle{ieeenat_fullname}
  \bibliography{main}
}

\appendix
\def\issupplement{1}
\renewcommand{\thesection}{S\arabic{section}}
\renewcommand{\thesubsection}{S\arabic{section}.\arabic{subsection}}
\renewcommand{\thefigure}{S\arabic{figure}}
\renewcommand{\thetable}{S\arabic{table}}
\renewcommand{\theequation}{S\arabic{equation}}
\setcounter{section}{0}
\setcounter{figure}{0}
\setcounter{table}{0}
\setcounter{equation}{0}

\begin{center}
{\Large\bf Supplementary Material}
\end{center}
\suppressfloats[t]

This supplementary reports the extended comparisons, random-seed
analysis, out-of-domain evaluation, backbone transfer, frame-level
localization, ablations, efficiency, adaptive-budget analysis, and
per-video routing study. Main-paper Fig.~\ref{fig:qual_failures} visualizes
the characteristic failure modes, and Algorithm~\ref{alg:ic} gives the
complete Information Compression procedure.

\begin{figure*}[t]
\centering
\begin{subfigure}{\linewidth}
\centering
\includegraphics[width=\linewidth]{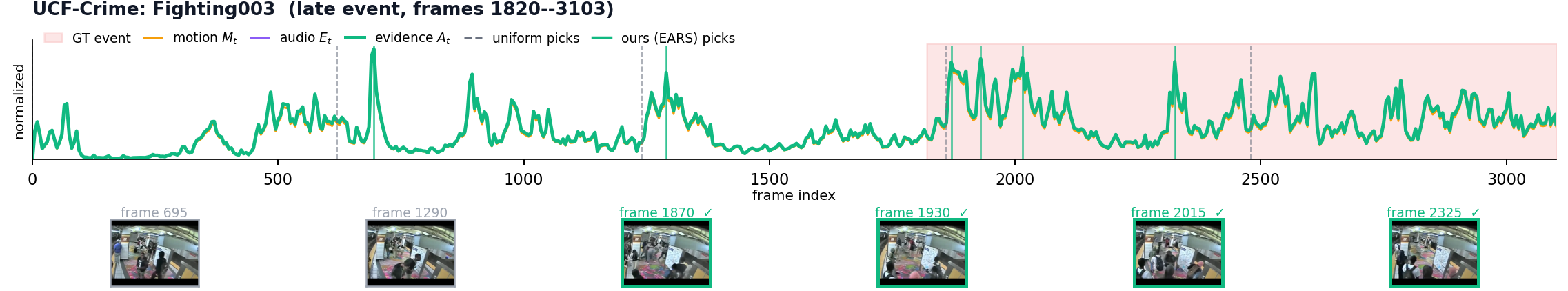}
\caption{UCF-Crime \emph{Fighting003}.}
\label{fig:supp_qualitative_a}
\end{subfigure}
\vspace{2pt}
\begin{subfigure}{\linewidth}
\centering
\includegraphics[width=\linewidth]{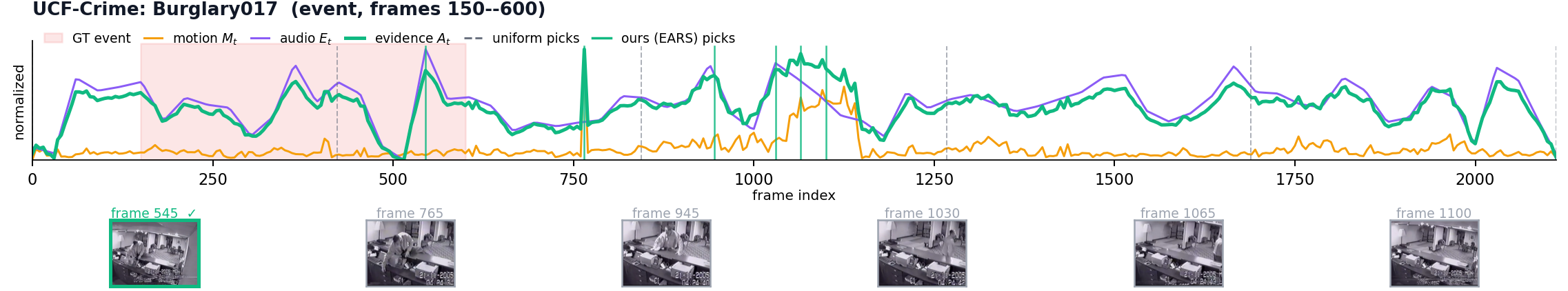}
\caption{UCF-Crime \emph{Burglary017}.}
\label{fig:supp_qualitative_b}
\end{subfigure}
\caption{\textbf{Motion and audio evidence on real UCF-Crime videos.}
For each example we show normalized motion $M_t$ (orange), audio energy
$E_t$ (purple), and fused evidence $A_t$ (green). The red band marks
the ground-truth interval, gray dashed lines are Uniform picks, and
green solid lines are EARS picks after IC. Below each plot are the six
EARS-selected frames; green borders mark frames inside the event.}
\label{fig:supp_qualitative}
\end{figure*}

\section{Frame-Level Curves}
\begin{figure}[H]
\centering
\includegraphics[width=\linewidth]{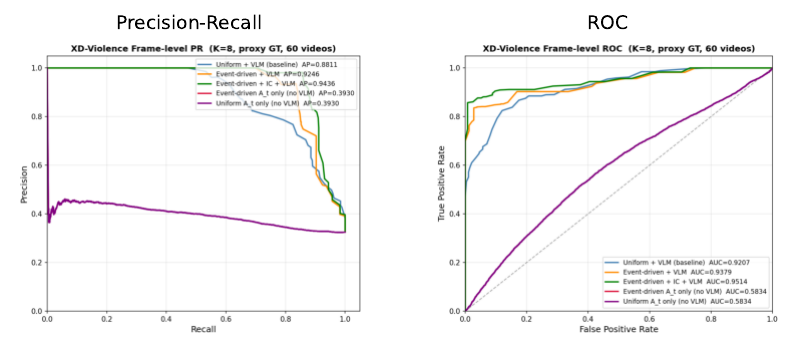}
\caption{Precision--Recall (left) and ROC (right) curves on the
frame-level XD-Violence annotations.}
\label{fig:supp_framelevel_curves}
\end{figure}

\section{Updated Matched Controls}
\label{sec:updated_protocol}

This new study is separate from the original tables and figures retained
above. It corrects video identities/labels and the Uniform baseline,
and uses the updated implementation described below. Its numbers must
not be mixed with the earlier protocols or interpreted as revalidation
of their localization, backbone-transfer, held-out, or routing claims.
Table~\ref{tab:main_validation} preserves the original main-paper
summary relocated here for space; its numbers are not new controls.

\begin{table}[H]
\centering
\caption{Original robustness, scope, and efficiency summary, relocated
from the main paper without changing its numerical findings. Values
are F1 unless noted; latency is measured before cache reuse.}
\label{tab:main_validation}
\footnotesize
\setlength{\tabcolsep}{3.5pt}
\begin{tabular}{p{0.22\columnwidth}p{0.72\columnwidth}}
\toprule
Study & Quantitative finding \\
\midrule
Flow estimator & TV-L1~\cite{tvl1} stays within $0.024$ F1 of Farneback~\cite{farneback2003} for
XD at every $K\leq4$. \\
Dense narrative & EgoSchema~\cite{egoschema} EARS$-$Uniform accuracy is
$-0.056/-0.090$ at $K=4/8$. \\
Per-video routing & Modality-statistics router $0.635$; always-EARS $0.685$;
oracle upper bound $0.789$. \\
End-to-end cost & Paired $n\!=\!24$: EARS $6.87{\pm}2.95$ vs.
Uniform $1.33{\pm}0.53$ s/video; reusable CPU preprocessing dominates. \\
\bottomrule
\end{tabular}
\end{table}

\subsection{Cohorts, Reasoner, and Signal Alignment}

XD comprises 120 Fighting, 84 Shooting, and 300 Normal videos in the
available local test subset, not the full official test benchmark.
UCF comprises 140 anomalous and 150 normal unique test videos;
duplicate entries and conflicting labels from the older loader are
not retained. Median nominal durations are 99.52 and 59.62 seconds.
Of the UCF signals, 222 are identically zero in audio amplitude;
zero amplitude does not itself prove stream absence.

All selectors use Qwen2-VL-7B-Instruct~\cite{qwen2vl}, revision
\nolinkurl{eed13092ef92e448dd6875b2a00151bd3f7db0ac}, bfloat16,
seed 42, greedy generation, and at most 16 new tokens.
Selected RGB frames are chronologically ordered and resized with
aspect-preserving \texttt{INTER\_AREA} interpolation to maximum side 448.
The reasoner sees task text and RGB, not audio, scores, flow, or timestamps.
Within each cohort every selector uses exactly the same binary prompt.

\paragraph{XD prompt.}
\begin{quote}\small
Given the following frames sampled from a surveillance video, determine
whether the video contains violent activity. Answer Yes or No.
\end{quote}

\paragraph{UCF prompt.}
\begin{quote}\small
These frames are from a surveillance video. The video may show: fighting,
robbery, abuse, road accidents, burglary, vandalism, explosions, or other
crimes. If any anomalous or criminal event is visible in these frames,
answer Yes. If the frames show only normal, safe activity, answer No.
Answer with one word: Yes or No.
\end{quote}

Uniform uses equal-bin centers, so $K=1$ is the midpoint.
Random uses three frozen without-replacement streams over native frames;
its sample SD is not a confidence interval for a method difference.
Farneb\"ack flow uses up to 256 endpoint-inclusive full-video candidates
in quarter-scale grayscale. The first motion value is zero; displacement
magnitudes are interpolated at actual native frame positions, not
divided by candidate time gaps.
Mono audio at 22,050 Hz is summarized by one-second mean absolute
amplitude. Nominal window centers are $j+0.5$ seconds, including a final
partial window; interpolation at frame timestamps uses constant endpoint
extension. Constant signals normalize to zero.
Fusion uses $\sigma(\tilde{E}_t+\tilde{M}_t)$ and native-frame NMS
radius $\operatorname{round}(0.5\,\mathrm{fps})$, with earliest ties.
Flat signals do not fall back to Uniform.
Every saved selection contains exactly $K$ unique in-range indices.
Only verified absent audio is zero-filled; probing, decoding,
invalid answers, or infeasible selections stop evaluation.

\subsection{Semantic Fusion and Simple Alternatives}

Semantic scoring uses frozen SigLIP-base-patch16-224~\cite{siglip},
revision \nolinkurl{7fd15f0689c79d79e38b1c2e2e2370a7bf2761ed}.
Unit-normalized image/text embeddings give cosine scores maximized
over eight fixed descriptions per cohort, not chosen from video labels:
\begin{itemize}
\setlength{\itemsep}{0pt}
\item XD: a violent scene; people fighting; a shooting; an explosion;
a car accident; an assault; abuse; a riot.
\item UCF: a crime being committed; an anomalous event; fighting or
assault; robbery or burglary; arson or a fire; a road accident;
shoplifting or stealing; vandalism.
\end{itemize}
Plain semantic top-$K$ ranks up to 256 full-video candidates.
Semantic + NMS and the hybrid use the same candidate-restricted NMS.
The hybrid averages independently candidate-normalized semantic and
sigmoid EARS scores at fixed weight $1/2$, without outcome-based tuning.
These are explicit controls, not an official AKS reproduction.
Comparing the hybrid to dense-grid EARS changes candidate support too.

Difference + audio replaces flow with mean absolute floating-point
grayscale difference divided by 255 between the same candidate frames.
The first value is zero; interpolation, audio, normalization, fusion,
and native-grid NMS otherwise match EARS.

\begin{table*}[t]
\centering\small
\caption{Updated matched F1, all methods and budgets. Random is mean
$\pm$ sample SD over three streams. The upper/lower groups distinguish
selection without/with a learned semantic encoder and task query.}
\label{tab:updated_all_f1}
\setlength{\tabcolsep}{6pt}
\begin{tabular}{lrrrrrr}
\toprule
& \multicolumn{3}{c}{XD subset ($n=504$)} & \multicolumn{3}{c}{UCF-Crime ($n=290$)} \\
\cmidrule(lr){2-4}\cmidrule(lr){5-7}
Selector & $K=1$ & $K=2$ & $K=4$ & $K=1$ & $K=2$ & $K=4$ \\
\midrule
\multicolumn{7}{l}{\textit{Query-independent selection; no learned selection encoder}} \\
Uniform & 0.614 & 0.791 & 0.893 & 0.303 & 0.528 & 0.713 \\
Random (3 seeds) & $0.603\pm0.010$ & $0.779\pm0.005$ & $0.853\pm0.015$ & $0.249\pm0.030$ & $0.537\pm0.028$ & $0.668\pm0.019$ \\
Motion & 0.524 & 0.743 & 0.856 & 0.270 & 0.471 & 0.685 \\
Audio & 0.683 & 0.838 & 0.907 & 0.095 & 0.158 & 0.272 \\
Difference + audio & 0.612 & 0.813 & 0.898 & 0.237 & 0.437 & 0.642 \\
EARS & 0.660 & 0.817 & 0.898 & 0.248 & 0.471 & 0.654 \\
\midrule
\multicolumn{7}{l}{\textit{Task-aware selection; frozen SigLIP encoder}} \\
Semantic top-$K$ & 0.824 & 0.843 & 0.832 & 0.546 & 0.651 & 0.694 \\
Semantic + NMS & 0.824 & 0.845 & 0.847 & 0.546 & 0.679 & 0.730 \\
Semantic + EARS & 0.840 & 0.868 & 0.889 & 0.351 & 0.651 & 0.739 \\
\bottomrule
\end{tabular}

\end{table*}

Table~\ref{tab:updated_all_f1} retains the complete comparison, including
Audio's higher XD point estimates and the hybrid's UCF one-frame loss.
All six EARS-minus-Uniform intervals include zero.
Semantic NMS alone improves four-frame F1 over plain top-$K$ by
$0.016$ on XD and $0.037$ on UCF, with exploratory intervals
$[0.001,0.033]$ and $[0.009,0.069]$.
That change must not be attributed to activity fusion.
All six Difference-minus-EARS intervals include zero, establishing
neither flow superiority nor proxy equivalence.

\begin{table*}[t]
\centering\small
\caption{Updated XD subset: complete primary metrics and confusion
counts, including all three Random streams. Every row covers 504 videos.}
\label{tab:updated_xd_metrics}
\begin{tabular}{lrrrrrrrrr}
\toprule
Selector & $K$ & Acc. & Prec. & Recall & F1 & TP & FP & FN & TN \\
\midrule
Uniform & 1 & 0.768 & 0.939 & 0.456 & 0.614 & 93 & 6 & 111 & 294 \\
Uniform & 2 & 0.851 & 0.916 & 0.696 & 0.791 & 142 & 13 & 62 & 287 \\
Uniform & 4 & 0.917 & 0.931 & 0.858 & 0.893 & 175 & 13 & 29 & 287 \\
EARS & 1 & 0.786 & 0.921 & 0.515 & 0.660 & 105 & 9 & 99 & 291 \\
EARS & 2 & 0.867 & 0.920 & 0.735 & 0.817 & 150 & 13 & 54 & 287 \\
EARS & 4 & 0.919 & 0.914 & 0.882 & 0.898 & 180 & 17 & 24 & 283 \\
Random-0 & 1 & 0.762 & 0.957 & 0.431 & 0.595 & 88 & 4 & 116 & 296 \\
Random-0 & 2 & 0.841 & 0.913 & 0.672 & 0.774 & 137 & 13 & 67 & 287 \\
Random-0 & 4 & 0.897 & 0.909 & 0.828 & 0.867 & 169 & 17 & 35 & 283 \\
Random-1 & 1 & 0.764 & 0.947 & 0.441 & 0.602 & 90 & 5 & 114 & 295 \\
Random-1 & 2 & 0.849 & 0.932 & 0.676 & 0.784 & 138 & 10 & 66 & 290 \\
Random-1 & 4 & 0.877 & 0.908 & 0.775 & 0.836 & 158 & 16 & 46 & 284 \\
Random-2 & 1 & 0.768 & 0.939 & 0.456 & 0.614 & 93 & 6 & 111 & 294 \\
Random-2 & 2 & 0.845 & 0.926 & 0.672 & 0.778 & 137 & 11 & 67 & 289 \\
Random-2 & 4 & 0.887 & 0.889 & 0.824 & 0.855 & 168 & 21 & 36 & 279 \\
Motion & 1 & 0.730 & 0.915 & 0.368 & 0.524 & 75 & 7 & 129 & 293 \\
Motion & 2 & 0.825 & 0.920 & 0.623 & 0.743 & 127 & 11 & 77 & 289 \\
Motion & 4 & 0.891 & 0.921 & 0.799 & 0.856 & 163 & 14 & 41 & 286 \\
Audio & 1 & 0.796 & 0.917 & 0.544 & 0.683 & 111 & 10 & 93 & 290 \\
Audio & 2 & 0.881 & 0.934 & 0.760 & 0.838 & 155 & 11 & 49 & 289 \\
Audio & 4 & 0.925 & 0.907 & 0.907 & 0.907 & 185 & 19 & 19 & 281 \\
\bottomrule
\end{tabular}

\end{table*}

\begin{table*}[t]
\centering\small
\caption{Updated UCF: complete primary metrics and confusion counts.
Every row covers 290 unique videos with annotation-derived labels.}
\label{tab:updated_ucf_metrics}
\begin{tabular}{lrrrrrrrrr}
\toprule
Selector & $K$ & Acc. & Prec. & Recall & F1 & TP & FP & FN & TN \\
\midrule
Uniform & 1 & 0.603 & 1.000 & 0.179 & 0.303 & 25 & 0 & 115 & 150 \\
Uniform & 2 & 0.679 & 0.912 & 0.371 & 0.528 & 52 & 5 & 88 & 145 \\
Uniform & 4 & 0.772 & 0.911 & 0.586 & 0.713 & 82 & 8 & 58 & 142 \\
EARS & 1 & 0.583 & 0.952 & 0.143 & 0.248 & 20 & 1 & 120 & 149 \\
EARS & 2 & 0.659 & 0.936 & 0.314 & 0.471 & 44 & 3 & 96 & 147 \\
EARS & 4 & 0.741 & 0.922 & 0.507 & 0.654 & 71 & 6 & 69 & 144 \\
Random-0 & 1 & 0.593 & 1.000 & 0.157 & 0.272 & 22 & 0 & 118 & 150 \\
Random-0 & 2 & 0.683 & 0.944 & 0.364 & 0.526 & 51 & 3 & 89 & 147 \\
Random-0 & 4 & 0.755 & 0.960 & 0.514 & 0.670 & 72 & 3 & 68 & 147 \\
Random-1 & 1 & 0.590 & 1.000 & 0.150 & 0.261 & 21 & 0 & 119 & 150 \\
Random-1 & 2 & 0.683 & 0.980 & 0.350 & 0.516 & 49 & 1 & 91 & 149 \\
Random-1 & 4 & 0.766 & 0.974 & 0.529 & 0.685 & 74 & 2 & 66 & 148 \\
Random-2 & 1 & 0.572 & 0.944 & 0.121 & 0.215 & 17 & 1 & 123 & 149 \\
Random-2 & 2 & 0.707 & 0.982 & 0.400 & 0.569 & 56 & 1 & 84 & 149 \\
Random-2 & 4 & 0.738 & 0.921 & 0.500 & 0.648 & 70 & 6 & 70 & 144 \\
Motion & 1 & 0.590 & 0.957 & 0.157 & 0.270 & 22 & 1 & 118 & 149 \\
Motion & 2 & 0.659 & 0.936 & 0.314 & 0.471 & 44 & 3 & 96 & 147 \\
Motion & 4 & 0.762 & 0.949 & 0.536 & 0.685 & 75 & 4 & 65 & 146 \\
Audio & 1 & 0.541 & 1.000 & 0.050 & 0.095 & 7 & 0 & 133 & 150 \\
Audio & 2 & 0.559 & 1.000 & 0.086 & 0.158 & 12 & 0 & 128 & 150 \\
Audio & 4 & 0.593 & 1.000 & 0.157 & 0.272 & 22 & 0 & 118 & 150 \\
\bottomrule
\end{tabular}

\end{table*}

\begin{table*}[t]
\centering\footnotesize
\caption{Updated semantic and targeted-control metrics. Same cohorts,
reasoner, binary prompts, and budgets as the primary controls.}
\label{tab:updated_new_metrics}
\setlength{\tabcolsep}{4pt}
\begin{tabular}{llrrrrrrrrr}
\toprule
Cohort & Selector & $K$ & Acc. & Prec. & Recall & F1 & TP & FP & FN & TN \\
\midrule
XD subset & Semantic top-$K$ & 1 & 0.857 & 0.824 & 0.824 & 0.824 & 168 & 36 & 36 & 264 \\
XD subset & Semantic top-$K$ & 2 & 0.865 & 0.796 & 0.897 & 0.843 & 183 & 47 & 21 & 253 \\
XD subset & Semantic top-$K$ & 4 & 0.847 & 0.751 & 0.931 & 0.832 & 190 & 63 & 14 & 237 \\
XD subset & Semantic + NMS & 1 & 0.857 & 0.824 & 0.824 & 0.824 & 168 & 36 & 36 & 264 \\
XD subset & Semantic + NMS & 2 & 0.865 & 0.788 & 0.912 & 0.845 & 186 & 50 & 18 & 250 \\
XD subset & Semantic + NMS & 4 & 0.859 & 0.755 & 0.966 & 0.847 & 197 & 64 & 7 & 236 \\
XD subset & Semantic + EARS & 1 & 0.877 & 0.886 & 0.799 & 0.840 & 163 & 21 & 41 & 279 \\
XD subset & Semantic + EARS & 2 & 0.887 & 0.824 & 0.917 & 0.868 & 187 & 40 & 17 & 260 \\
XD subset & Semantic + EARS & 4 & 0.901 & 0.810 & 0.985 & 0.889 & 201 & 47 & 3 & 253 \\
XD subset & Difference + audio & 1 & 0.764 & 0.913 & 0.461 & 0.612 & 94 & 9 & 110 & 291 \\
XD subset & Difference + audio & 2 & 0.863 & 0.909 & 0.735 & 0.813 & 150 & 15 & 54 & 285 \\
XD subset & Difference + audio & 4 & 0.919 & 0.910 & 0.887 & 0.898 & 181 & 18 & 23 & 282 \\
UCF-Crime & Semantic top-$K$ & 1 & 0.697 & 0.981 & 0.379 & 0.546 & 53 & 1 & 87 & 149 \\
UCF-Crime & Semantic top-$K$ & 2 & 0.741 & 0.933 & 0.500 & 0.651 & 70 & 5 & 70 & 145 \\
UCF-Crime & Semantic top-$K$ & 4 & 0.766 & 0.939 & 0.550 & 0.694 & 77 & 5 & 63 & 145 \\
UCF-Crime & Semantic + NMS & 1 & 0.697 & 0.981 & 0.379 & 0.546 & 53 & 1 & 87 & 149 \\
UCF-Crime & Semantic + NMS & 2 & 0.755 & 0.926 & 0.536 & 0.679 & 75 & 6 & 65 & 144 \\
UCF-Crime & Semantic + NMS & 4 & 0.786 & 0.933 & 0.600 & 0.730 & 84 & 6 & 56 & 144 \\
UCF-Crime & Semantic + EARS & 1 & 0.617 & 0.968 & 0.214 & 0.351 & 30 & 1 & 110 & 149 \\
UCF-Crime & Semantic + EARS & 2 & 0.741 & 0.933 & 0.500 & 0.651 & 70 & 5 & 70 & 145 \\
UCF-Crime & Semantic + EARS & 4 & 0.786 & 0.898 & 0.629 & 0.739 & 88 & 10 & 52 & 140 \\
UCF-Crime & Difference + audio & 1 & 0.579 & 0.950 & 0.136 & 0.237 & 19 & 1 & 121 & 149 \\
UCF-Crime & Difference + audio & 2 & 0.645 & 0.930 & 0.286 & 0.437 & 40 & 3 & 100 & 147 \\
UCF-Crime & Difference + audio & 4 & 0.731 & 0.897 & 0.500 & 0.642 & 70 & 8 & 70 & 142 \\
\bottomrule
\end{tabular}

\end{table*}

\begin{table*}[t]
\centering\small
\caption{Updated EARS-minus-Uniform paired comparisons, all budgets.
Intervals use 10,000 joint video resamples, seed 42, and are unadjusted.}
\label{tab:updated_primary_paired}
\begin{tabular}{lrrlrrr}
\toprule
Cohort & $K$ & $\Delta$F1 & 95\% paired interval & Wins & Losses & Ties \\
\midrule
XD subset & 1 & $+0.047$ & $[-0.038,\ +0.132]$ & 58 & 49 & 397 \\
XD subset & 2 & $+0.026$ & $[-0.030,\ +0.085]$ & 50 & 42 & 412 \\
XD subset & 4 & $+0.005$ & $[-0.033,\ +0.044]$ & 30 & 29 & 445 \\
UCF-Crime & 1 & $-0.055$ & $[-0.166,\ +0.060]$ & 13 & 19 & 258 \\
UCF-Crime & 2 & $-0.057$ & $[-0.168,\ +0.055]$ & 27 & 33 & 230 \\
UCF-Crime & 4 & $-0.059$ & $[-0.134,\ +0.018]$ & 20 & 29 & 241 \\
\bottomrule
\end{tabular}

\end{table*}

\begin{table*}[t]
\centering\small
\caption{All 36 updated targeted-control contrasts, including
frame differences versus EARS, NMS-only effects, and activity fusion.
Exploratory unadjusted 95\% paired intervals, not confirmatory tests.}
\label{tab:updated_controls_paired}
\setlength{\tabcolsep}{4pt}
\begin{tabular}{lllrll}
\toprule
Cohort & Selector & Reference & $K$ & $\Delta$F1 & 95\% interval \\
\midrule
UCF-Crime & Semantic + NMS & Semantic top-$K$ & 1 & $+0.000$ & $[+0.000, +0.000]$ \\
UCF-Crime & Semantic + NMS & Semantic top-$K$ & 2 & $+0.028$ & $[-0.003, +0.061]$ \\
UCF-Crime & Semantic + NMS & Semantic top-$K$ & 4 & $+0.037$ & $[+0.009, +0.069]$ \\
UCF-Crime & Semantic + EARS & Semantic + NMS & 1 & $-0.196$ & $[-0.302, -0.093]$ \\
UCF-Crime & Semantic + EARS & Semantic + NMS & 2 & $-0.028$ & $[-0.113, +0.058]$ \\
UCF-Crime & Semantic + EARS & Semantic + NMS & 4 & $+0.009$ & $[-0.048, +0.066]$ \\
UCF-Crime & Semantic + EARS & Semantic top-$K$ & 1 & $-0.196$ & $[-0.302, -0.093]$ \\
UCF-Crime & Semantic + EARS & Semantic top-$K$ & 2 & $+0.000$ & $[-0.086, +0.088]$ \\
UCF-Crime & Semantic + EARS & Semantic top-$K$ & 4 & $+0.046$ & $[-0.018, +0.111]$ \\
UCF-Crime & Semantic + EARS & EARS & 1 & $+0.102$ & $[+0.026, +0.181]$ \\
UCF-Crime & Semantic + EARS & EARS & 2 & $+0.181$ & $[+0.100, +0.267]$ \\
UCF-Crime & Semantic + EARS & EARS & 4 & $+0.085$ & $[+0.025, +0.149]$ \\
UCF-Crime & Difference + audio & EARS & 1 & $-0.011$ & $[-0.094, +0.071]$ \\
UCF-Crime & Difference + audio & EARS & 2 & $-0.033$ & $[-0.125, +0.057]$ \\
UCF-Crime & Difference + audio & EARS & 4 & $-0.012$ & $[-0.085, +0.057]$ \\
UCF-Crime & Difference + audio & Uniform & 1 & $-0.066$ & $[-0.173, +0.047]$ \\
UCF-Crime & Difference + audio & Uniform & 2 & $-0.091$ & $[-0.205, +0.026]$ \\
UCF-Crime & Difference + audio & Uniform & 4 & $-0.071$ & $[-0.155, +0.010]$ \\
XD subset & Semantic + NMS & Semantic top-$K$ & 1 & $+0.000$ & $[+0.000, +0.000]$ \\
XD subset & Semantic + NMS & Semantic top-$K$ & 2 & $+0.002$ & $[-0.012, +0.017]$ \\
XD subset & Semantic + NMS & Semantic top-$K$ & 4 & $+0.016$ & $[+0.001, +0.033]$ \\
XD subset & Semantic + EARS & Semantic + NMS & 1 & $+0.017$ & $[-0.023, +0.057]$ \\
XD subset & Semantic + EARS & Semantic + NMS & 2 & $+0.022$ & $[-0.009, +0.054]$ \\
XD subset & Semantic + EARS & Semantic + NMS & 4 & $+0.042$ & $[+0.017, +0.067]$ \\
XD subset & Semantic + EARS & Semantic top-$K$ & 1 & $+0.017$ & $[-0.023, +0.057]$ \\
XD subset & Semantic + EARS & Semantic top-$K$ & 2 & $+0.024$ & $[-0.009, +0.058]$ \\
XD subset & Semantic + EARS & Semantic top-$K$ & 4 & $+0.058$ & $[+0.031, +0.087]$ \\
XD subset & Semantic + EARS & EARS & 1 & $+0.180$ & $[+0.117, +0.245]$ \\
XD subset & Semantic + EARS & EARS & 2 & $+0.050$ & $[-0.000, +0.100]$ \\
XD subset & Semantic + EARS & EARS & 4 & $-0.008$ & $[-0.044, +0.027]$ \\
XD subset & Difference + audio & EARS & 1 & $-0.048$ & $[-0.126, +0.027]$ \\
XD subset & Difference + audio & EARS & 2 & $-0.004$ & $[-0.051, +0.043]$ \\
XD subset & Difference + audio & EARS & 4 & $+0.001$ & $[-0.030, +0.031]$ \\
XD subset & Difference + audio & Uniform & 1 & $-0.001$ & $[-0.087, +0.082]$ \\
XD subset & Difference + audio & Uniform & 2 & $+0.022$ & $[-0.038, +0.082]$ \\
XD subset & Difference + audio & Uniform & 4 & $+0.005$ & $[-0.032, +0.044]$ \\
\bottomrule
\end{tabular}

\end{table*}

\subsection{Paired Cost and Cache Reuse}

At $K=4$, 24 videos (12 per cohort) are stratified by label and
within-label duration tertile, with two seeded draws per stratum.
Uniform, EARS, and semantic top-$K$ each receive three measured
repetitions under fresh-feature and cached-feature conditions:
432 fresh-answer trials, excluding 72 warm-ups.
On NVIDIA GB10 the models remain resident, CUDA stages are synchronized,
and the OS file cache is explicitly warmed. Model loading and cold-disk
access are excluded. Fresh-feature trials include extraction and cache
creation; cached semantic selection still encodes its queries.
Every condition then uses a common selected-RGB decode/resize path
and a newly inferred VLM answer. A seeded cyclic order balances trials.

We aggregate repetitions within video before comparing videos.
Marginal stage medians need not add to median total time.
Paired differences are medians of per-video differences, not differences
of marginal medians; repetitions are not independent bootstrap samples.

\begin{table*}[t]
\centering\small
\caption{Updated cost stages in seconds: medians over per-video
repetition medians. Fresh/cached refer to video features, not a
cold/warm model or OS file cache.}
\label{tab:updated_cost_stages}
\setlength{\tabcolsep}{4pt}
\begin{tabular}{lllrrrr}
\toprule
Cohort & Selector & Features & Selection & RGB & VLM & Total \\
\midrule
XD subset & Uniform & Fresh & 0.0002 & 0.2281 & 0.5215 & 0.7657 \\
UCF-Crime & Uniform & Fresh & 0.0002 & 0.1314 & 0.4664 & 0.6009 \\
XD subset & Uniform & Cached & 0.0002 & 0.2244 & 0.5222 & 0.7623 \\
UCF-Crime & Uniform & Cached & 0.0002 & 0.1318 & 0.4707 & 0.5973 \\
XD subset & EARS & Fresh & 0.7656 & 0.0971 & 0.5262 & 1.3775 \\
UCF-Crime & EARS & Fresh & 0.3958 & 0.1263 & 0.4685 & 0.9773 \\
XD subset & EARS & Cached & 0.0007 & 0.0950 & 0.5240 & 0.6111 \\
UCF-Crime & EARS & Cached & 0.0007 & 0.1248 & 0.4693 & 0.5926 \\
XD subset & Semantic top-$K$ & Fresh & 0.9889 & 0.1743 & 0.5219 & 1.7525 \\
UCF-Crime & Semantic top-$K$ & Fresh & 0.6690 & 0.1023 & 0.4705 & 1.2567 \\
XD subset & Semantic top-$K$ & Cached & 0.0116 & 0.1764 & 0.5223 & 0.7223 \\
UCF-Crime & Semantic top-$K$ & Cached & 0.0118 & 0.1024 & 0.4693 & 0.5792 \\
\bottomrule
\end{tabular}

\end{table*}

\begin{table*}[t]
\centering\small
\caption{All updated paired total-cost comparisons with Uniform.
Video-bootstrap intervals and median paired total ratios; positive
differences or ratios above one indicate greater cost.}
\label{tab:updated_cost_paired}
\setlength{\tabcolsep}{4pt}
\begin{tabular}{lllrlr}
\toprule
Cohort & Selector & Features & $\Delta$ seconds & 95\% interval & Ratio \\
\midrule
UCF-Crime & EARS & Fresh & $+0.3856$ & $[+0.2739, +0.6267]$ & 1.652 \\
UCF-Crime & EARS & Cached & $-0.0079$ & $[-0.0146, +0.0131]$ & 0.987 \\
UCF-Crime & Semantic top-$K$ & Fresh & $+0.6239$ & $[+0.5952, +0.7634]$ & 2.084 \\
UCF-Crime & Semantic top-$K$ & Cached & $-0.0255$ & $[-0.1512, -0.0030]$ & 0.956 \\
XD subset & EARS & Fresh & $+0.6470$ & $[+0.5045, +1.3221]$ & 1.967 \\
XD subset & EARS & Cached & $-0.0278$ & $[-0.1079, +0.0071]$ & 0.954 \\
XD subset & Semantic top-$K$ & Fresh & $+0.9138$ & $[+0.7152, +1.1870]$ & 2.226 \\
XD subset & Semantic top-$K$ & Cached & $+0.0057$ & $[-0.0158, +0.0400]$ & 1.007 \\
\bottomrule
\end{tabular}

\end{table*}

Fresh-feature EARS has lower marginal medians than semantic selection
but higher total cost than Uniform. Both cached EARS-minus-Uniform
total-time intervals include zero, so no reliable cached speedup or
equivalence is established. The hybrid, semantic-NMS, and difference
selectors have no matched timing study.

The primary, semantic, and targeted runs contain 16,674, 2,382,
and 7,146 classification records, respectively, covering 66 complete
cohort/method/budget cells. Audits check identities, labels, strict
raw-answer parsing, selections, confusion counts, frozen sources,
model revisions, and signal/feature hashes. Deterministic reuse of
identical-input answers in quality evaluation is logged and is never
used as timing evidence.

\end{document}